\documentclass{article} % For LaTeX2e
\usepackage[final]{colm2026_conference}
\usepackage[utf8]{inputenc} % allow utf-8 input
\usepackage[T1]{fontenc}    % use 8-bit T1 fonts
\usepackage{hyperref}       % hyperlinks
\usepackage{url}            % simple URL typesetting
\usepackage{booktabs}       % professional-quality tables
\usepackage{amsfonts}       % blackboard math symbols
\usepackage{nicefrac}       % compact symbols for 1/2, etc.
\usepackage{microtype}      % microtypography
\usepackage{xcolor}         % colors
\usepackage[most]{tcolorbox}      % colored boxes

\usepackage{enumitem}
\usepackage{graphicx}
\usepackage{multirow}
\usepackage{colortbl}
\usepackage{amsmath} 
\usepackage{fontawesome5}
\usepackage{subfiles}
\usepackage{array}
\usepackage{makecell}
\usepackage{pifont}
\usepackage{wrapfig}
\usepackage{graphicx}
\usepackage{subcaption}
\usepackage{CJKutf8}

\usepackage{lineno}

\definecolor{darkblue}{rgb}{0, 0, 0.5}
\hypersetup{colorlinks=true, citecolor=darkblue, linkcolor=darkblue, urlcolor=darkblue}

\newcommand{\greentick}{\textcolor{teal}{\ding{52}}}
\newcommand{\redcross}{\textcolor{red}{\ding{56}}}
\newcommand{\question}{\scalebox{0.8}{\faQuestion}}
\newcommand{\errstat}[2]{\makecell[c]{#1\\(#2)}}

\DeclareRobustCommand{\faVolumeDowns}{%
  \protect\textcolor{black!40}{\protect\faVolumeDown}%
  \hspace{-0.6em}\raisebox{0.2em}{\protect\textcolor{black}{\protect\faVolumeDown}}%
}
\title{MMMG: A Comprehensive and Reliable Benchmark \\ for \underline{M}ultitask \underline{M}ulti\underline{m}odal \underline{G}eneration\\
}

\author{
    Jihan Yao\thanks{equal contribution} \ \textsuperscript{1}, 
    Yushi Hu\footnotemark[1] \ \textsuperscript{1}, 
    Wenyuan Wang\textsuperscript{1},
    Bin Han\textsuperscript{1},
    Shangbin Feng\textsuperscript{1}, 
    Guang Yang\textsuperscript{1} \\
    \textbf{Yujie Yi}\textsuperscript{1},
    \textbf{Bingbing Wen}\textsuperscript{1},
    \textbf{Ranjay Krishna}\textsuperscript{1,}\textsuperscript{2}, 
    \textbf{Lucy Lu Wang}\textsuperscript{1,}\textsuperscript{2}, 
    \textbf{Yulia Tsvetkov}\textsuperscript{1} \\
    \textbf{Noah A. Smith}\textsuperscript{1,}\textsuperscript{2}, 
    \textbf{Banghua Zhu}\textsuperscript{1}  \\
    \textsuperscript{1}University of Washington  \ \ \ 
    \textsuperscript{2}Allen Institute for AI \\
    \texttt{jihany2@cs.washington.edu,} \quad
    \texttt{yushihu@uw.edu}
}

\begin{document}

\ifcolmsubmission
\linenumbers
\fi

\maketitle

\begin{abstract}
% \ranjay{Set up the problem first. Why is there a need for a benchmark like this?}
Automatically evaluating multimodal generation presents a significant challenge, as automated metrics often struggle to align with human evaluation, especially for complex tasks that involve multiple modalities. We present \textsc{MMMG}, the first benchmark to bring the verifiable-task paradigm to multimodal generation, spanning 4 modality combinations (image, audio, interleaved text and image, interleaved text and audio). As few multimodal outputs can be checked by programs alone, \textsc{MMMG} targets tasks that are either \emph{verifiable} or \emph{near-verifiable}: by providing references, and constraining model judges with explicit rubrics. We keep tasks challenging for generation models while enabling reliable automatic evaluation through a combination of models and programs. \textsc{MMMG} encompasses 55 tasks (including 31 newly developed ones), each with a carefully designed evaluation pipeline, and 1288 instructions to systematically assess reasoning, controllability, and other key capabilities of multimodal generation models. Extensive validation demonstrates that \textsc{MMMG} is highly aligned with human judgment, achieving an average agreement of 94.4\%. Benchmarking results on 29 models reveal that even though the state-of-the-art model, \textsc{GPT Image}, achieves 70.7\% accuracy for image generation, it falls short on interleaved generation. Furthermore, results suggest considerable improvement space in audio generation, highlighting an important future direction. 
% Code and data are publicly available at \href{https://github.com/yaojh18/MMMG}{https://github.com/yaojh18/MMMG}.

\end{abstract}

\section{Introduction}
\begin{figure}[!t]
    \centering
    \includegraphics[width=1\linewidth]{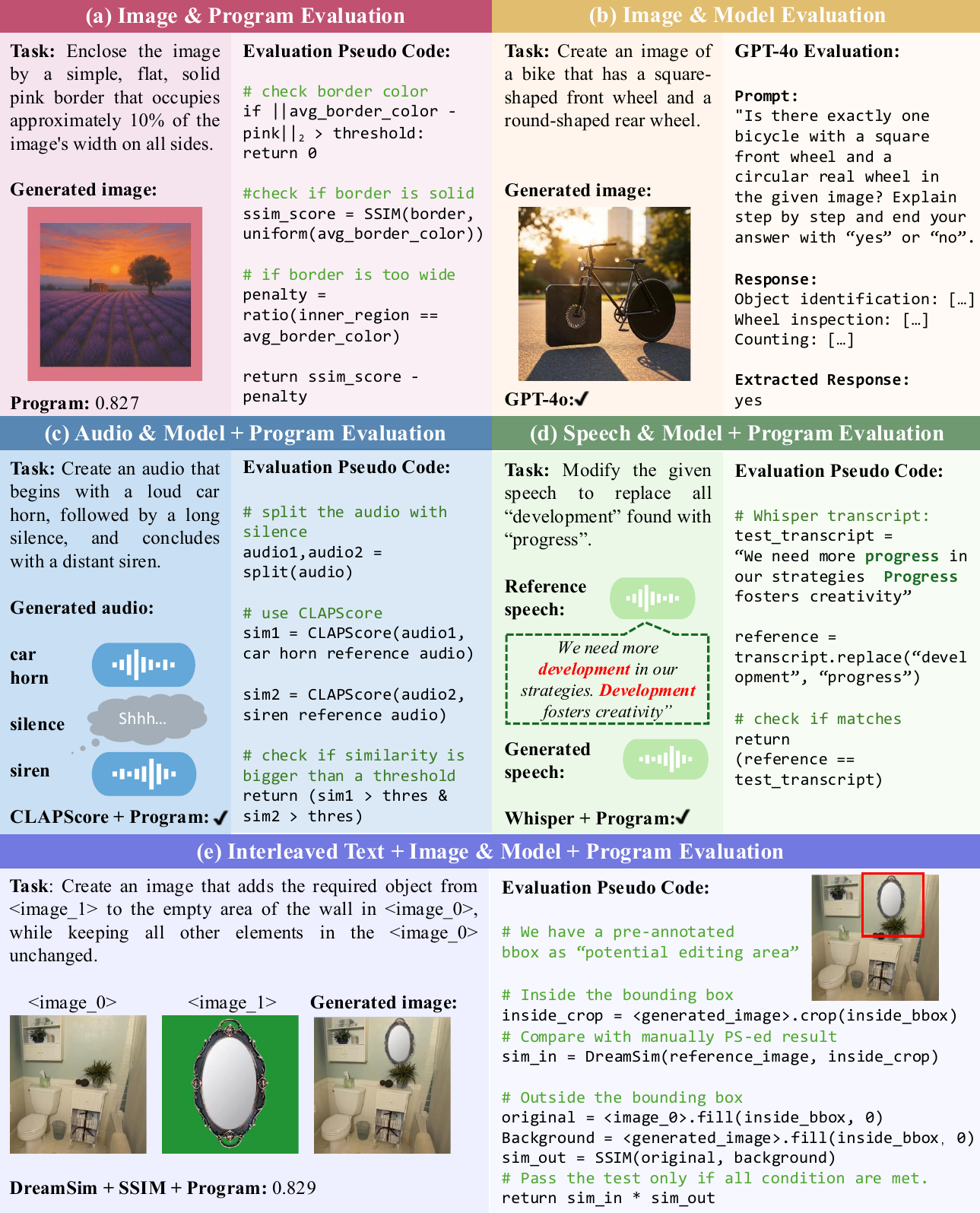}
    \caption{Examples of tasks and their evaluation in \textsc{MMMG}. For each task, we develop an evaluation metric using programs, models or their combinations. The tasks are either verifiable by programs or near-verifiable with big generation-evaluation gaps: generation is challenging, while strengthened task conditions allow automatic evaluations to easily align with human judgment. We show pseudo-code to demonstrate the evaluation process.}
    \label{fig:example}
    \vspace{-10pt}
\end{figure}
\begin{table}[!ht]\centering
    \scriptsize
    \tabcolsep=0.02cm
    \renewcommand*{\arraystretch}{1.2}
    \begin{tabular}{cllcccc}\toprule[1.5pt]
        \textbf{Task} & \multicolumn{1}{c}{\textbf{Subtask}} & \multicolumn{1}{c}{\textbf{Description}} & \multicolumn{1}{c}{\textbf{In.}} & \multicolumn{1}{c}{\textbf{Out.}} & \multicolumn{1}{c}{\textbf{\# }} & \multicolumn{1}{c}{\textbf{Evaluation}} \\\midrule[0.75pt]
        
        \multirow{6}{*}{\makecell{Object\\Generation}} &Inclusion &Include one or two unrelated objects in the scene. &$\mathbb{T}$ &\faImage &40 &VLM \\ 
        &\cellcolor[HTML]{d4edda}Exclusion &Exclude one related object from the scene. &$\mathbb{T}$ &\faImage &40 &VLM \\
        &Count &Generate exactly N objects. &$\mathbb{T}$ &\faImage &40 &VLM \\
        &\cellcolor[HTML]{d4edda}Attribution & Generate an object with uncommon attributes. &$\mathbb{T}$ &\faImage &40 &VLM \\
        &\cellcolor[HTML]{d4edda}Knowledge & Reason the answer object to a multi-hop question. &$\mathbb{T}$ &\faImage &40 &VLM \\
        &Commonsense & Reason the answer object/scene by commonsense. &$\mathbb{T}$ &\faImage &40 &VLM \\
        \hline 
        
        \multirow{4}{*}{\makecell{Relation\\Control}} &Comparison & Generate two objects with uncommon relations. &$\mathbb{T}$ &\faImage &20 &VLM \\
        &Universal & Generate objects with all identical/different attributes. &$\mathbb{T}$ &\faImage &20 &VLM \\      
        &Relative Spatial &Generate two objects with given relative spacial relation. &$\mathbb{T}$ &\faImage &20 &VLM \\
        &\cellcolor[HTML]{d4edda}Absolute Spatial &Generate one/two objects in asked absolute image quarter. &$\mathbb{T}$ &\faImage &20 &VLM \\ \hline 
        
        \multirow{2}{*}{\makecell{Image\\Format}} &\cellcolor[HTML]{d4edda}Border Fill &Surround the image with pure, solid and colored border. &$\mathbb{T}$ &\faImage &15 &Program + SSIM \\
        &\cellcolor[HTML]{d4edda}Region Fill &Fill the given region with pure and solid color. &$\mathbb{T}$ &\faImage &15 &Program + SSIM \\ \hline 
        
        \multirow{3}{*}{\makecell{Text\\Rendering}} &Single &Render English text on one object. &$\mathbb{T}$ &\faImage &20 &VLM \\
        &\cellcolor[HTML]{d4edda}Double &Render two English texts on two objects. &$\mathbb{T}$ &\faImage &20 &VLM \\
        &Multi-Lingual &Render one Chinese text on one object. &$\mathbb{T}$ &\faImage &20 &VLM \\ \hline 
        
        \multirow{8}{*}{\makecell{Image\\Editing}} &Object Adding &Add a new object by textual/visual prompt. &$\mathbb{T}$, \faImage &\faImage &20 &VLM + SSIM \\
        &Object Removing &Remove an existing object by textual/visual prompt. &$\mathbb{T}$, \faImage &\faImage &20 &VLM + SSIM \\
        &Object Replacing &Replace an existing object by textual/visual prompt. &$\mathbb{T}$, \faImage &\faImage &20 &VLM + SSIM \\
        &Object Altering &Change the attributes of objects by textual/visual prompt. &$\mathbb{T}$, \faImage &\faImage &20 &VLM + SSIM \\
        &Text Adding &Add text/long sentence by textual/visual prompt.&$\mathbb{T}$, \faImage &\faImage &20 &VLM + SSIM \\
        &\cellcolor[HTML]{d4edda}Text Altering &Remove/Modify/Translate text by textual/visual prompt.&$\mathbb{T}$, \faImage &\faImage &20 &VLM + SSIM \\
        &\cellcolor[HTML]{d4edda}Inter. Adding & Add an external image object to the original image. &$\mathbb{T}$, \faImages &\faImage &20 &DreamSim + SSIM \\
        &\cellcolor[HTML]{d4edda}Inter. Altering  &Change the color of an object with external color reference. &$\mathbb{T}$, \faImages &\faImage &20 &DreamSim + SSIM \\ \hline 
        
        \multirow{5}{*}{\makecell{Image\\Consistency}} &\cellcolor[HTML]{d4edda}Semantic &Generate multiple images in semantic order. &$\mathbb{T}$ &\faImages &20 &VLM \\
        &\cellcolor[HTML]{d4edda}Composition &Gradually add individual objects in the given order. &$\mathbb{T}$ &\faImages &20 &VLM \\
        &\cellcolor[HTML]{d4edda}Decomposition &Gradually remove object combination in the given order. &$\mathbb{T}$, \faImage &\faImages &20 &VLM \\
        &Multi-View &Generate multiple views of the reference scene. &$\mathbb{T}$, \faImage &\faImages &20 &SSIM \\
        &Multi-Angle & Generate multiple views of the reference object. &$\mathbb{T}$, \faImage &\faImages &20 &SSIM \\\hline 
        
        \multirow{6}{*}{\makecell{Image-Text\\Coherence}} &\cellcolor[HTML]{d4edda}Self Count & Count objects in the self-generated image. &$\mathbb{T}$ &$\mathbb{T}$, \faImage &20 &VLM \\
        &\cellcolor[HTML]{d4edda}Self Color & Name object colors in the self-generated image. &$\mathbb{T}$ &$\mathbb{T}$, \faImage &20 &VLM \\
        &\cellcolor[HTML]{d4edda}Self Size & Compare object sizes in the self-generated image. &$\mathbb{T}$ &$\mathbb{T}$, \faImage &20 &VLM \\
        &\cellcolor[HTML]{d4edda}Self Rel. Spatial & Decide relative spacial relation in the generated image.
        &$\mathbb{T}$ &$\mathbb{T}$, \faImage &20 &VLM \\
        &\cellcolor[HTML]{d4edda}Self Abs. Spatial & Decide absolute spacial relation in the generated image. &$\mathbb{T}$ &$\mathbb{T}$, \faImage &20 &VLM \\
        &\cellcolor[HTML]{d4edda}Self OCR &Recognize the text in the generated image. &$\mathbb{T}$ &$\mathbb{T}$, \faImage &20 &VLM \\\hline 
        
        \multirow{2}{*}{\makecell{Interleaved\\Reasoning}} &\cellcolor[HTML]{d4edda}Math &Solve the IQ-test puzzles. &$\mathbb{T}$, \faImage &$\mathbb{T}$, \faImage &20 &VLM \\
        &\cellcolor[HTML]{d4edda}Code &Generate/Debug and render frontend codes. &$\mathbb{T}$ &$\mathbb{T}$, \faImage &60 &VLM + DreamSim \\\hline 
        
        \multirow{4}{*}{\makecell{Sound\\Generation}} &Begin-End &Begin/End the audio with the given sound effect. &$\mathbb{T}$ &\faVolumeDown &20 &CLAPScore \\
        &\cellcolor[HTML]{d4edda}Positional Inclu. &Include one sound effect at a relative audio position.&$\mathbb{T}$ &\faVolumeDown &20 &CLAPScore \\
        &\cellcolor[HTML]{d4edda}Silence & Generate two ordered sound effects separated by silence.&$\mathbb{T}$ &\faVolumeDown &20 &CLAPScore \\
        &\cellcolor[HTML]{d4edda}Knowledge &Reason the answer sound to a multi-hop question. &$\mathbb{T}$ &\faVolumeDown &18 &CLAPScore \\\hline 
        
        \multirow{5}{*}{\makecell{Music\\Generation}} &Instrument Inclu. &Generate music with the given instrument. &$\mathbb{T}$ &\faVolumeDown &20 &CLAPScore \\
        &\cellcolor[HTML]{d4edda}Instrument Exclu. &Generate music without the given instrument. &$\mathbb{T}$ &\faVolumeDown &20 &CLAPScore \\
        &Genre &Generate music with the given genre. &$\mathbb{T}$ &\faVolumeDown &20 &CLAPScore \\
        &Tempo &Generate music with the given tempo. &$\mathbb{T}$ &\faVolumeDown &20 &Program \\
        &\cellcolor[HTML]{d4edda}Intensity &Generate music with fade in/out at the beginning/end. &$\mathbb{T}$ &\faVolumeDown &20 &Program \\ \hline 
        
        \multirow{8}{*}{\makecell{Interleaved\\Speech\\Generation}} &Voice Attribution & Generate an en. speech with required voice attributes. &$\mathbb{T}$ &\faVolumeDown &20 &\tiny{Whisper+W2V+Program}\\

        &Voice Replication &Generate an en. speech replicating the reference voice. &$\mathbb{T}$, \faVolumeDown &\faVolumeDown &20 &Whisper + WavLM \\

        &Multi-Lingual&Generate a zh. speech with required voice attributes. &$\mathbb{T}$ &\faVolumeDown &20 &\tiny{Whisper+W2V+Program} \\

        &Transcript Gen. &Generate a speech with textual constraints for transcripts. &$\mathbb{T}$ & \faVolumeDown &20 &Whisper + Program \\ 
        &\cellcolor[HTML]{d4edda}Transcript Edit &Editing a speech with textual constraints for transcripts. &$\mathbb{T}$, \faVolumeDown &\faVolumeDown &20 &Whisper + Program \\
        &\cellcolor[HTML]{d4edda}Conversation & Generate a conversation with given speaker order.&$\mathbb{T}$ &\faVolumeDowns &20 &Whisper + WavLM \\
        &Speech Translate & Directly translate multi-lingual speech into English speech.&$\mathbb{T}$, \faVolumeDown &\faVolumeDown &40 &Whisper + Program \\
        &\cellcolor[HTML]{d4edda}Speech Retrival & Retrieve the key speech segment in extremely long speech.&$\mathbb{T}$, \faVolumeDown &\faVolumeDown &40 &Whisper + Program \\
        \hline 
        
        \multirow{2}{*}{\makecell{Modality\\Order Control}} &\cellcolor[HTML]{d4edda}Image-Text &Generate interleaved image-text content in given order. &$\mathbb{T}$, \faImage &$\mathbb{T}$, \faImages &20 &Program \\
        &\cellcolor[HTML]{d4edda}Audio-Text &Generate interleaved audio-text content in given order. &$\mathbb{T}$, \faVolumeDown &$\mathbb{T}$, \faVolumeDowns &20 &Program \\
    \bottomrule[1.5pt]
    \end{tabular}
\vspace{5pt}
\caption{Detailed metadata for \textsc{MMMG}. $\mathbb{T}$ denotes text modality, \faImage \ for image modality, \faImages \ for multiple images, \faVolumeDown \ for audio and \faVolumeDowns \ for multiple audios. We evaluate each task with the method that most aligned with human judgment. \colorbox[HTML]{d4edda}{green background} indicates new tasks.}
\label{tab:detail}
\vspace{-10pt}
\end{table}
As investments in multimodal generative AI grow, current models are rapidly advancing their capabilities in generating text \citep{achiam2023gpt}, images \citep{podell2023sdxl}, audio \citep{evans2025stable}, and their interleaved content \citep{chen2025janus, wang2024emu3}. However, rigorous and reproducible evaluation of multimodal generation lags behind, raising a critical question: how can we accurately and effectively evaluate these models?

Human evaluations \citep{chiang2024chatbot, saharia2022photorealistic, liu2025musiceval}, while considered the gold standard, are prohibitively expensive for comprehensive assessment at scale. Moreover, inherent subjectivity makes it difficult to systematically identify specific model weaknesses. As an alternative, existing automated evaluation approaches face two main limitations. First, it is hard to align automatic evaluation metrics well with human judges. Most multimodal generation benchmarks~\citep{xia2024mmie, chen2024comm, chen2024interleaved} rely on multimodal language models as judges (MLM-as-a-judge)~\citep{hu2023tifa,chen2024mllm} without carefully validating their reliability, potentially causing misalignment with human judgment~\citep{chen2024mllm,pu2025judge}. Second, most benchmarks focus solely on single modalities~\citep{ji2024align, ghosh2023geneval, xie2025audiotime}, failing to capture the rich interleaved multimodal content (vision, language, speech/audio) that characterizes real-world tasks such as cross-modal reasoning \citep{hu2024visual}.

In text generation, \emph{verifiable} instructions, whose satisfaction can be objectively checked by programs \citep{zhou2023instruction}, have become a cornerstone of reliable evaluation and, more recently, of post-training with verifiable rewards \citep{lambert2024tulu, guo2025deepseek}. To address the above gaps, we introduce \textsc{MMMG}, to the best of our knowledge the first benchmark that extends this verifiable paradigm to multimodal generation. Strict programmatic verifiability, however, is rare beyond text: whether an image depicts a given scene rarely reduces to a simple rule. We therefore include tasks that meet one of two criteria: (1) \emph{verifiable} tasks as defined in IF-Eval \citep{zhou2023instruction}, where outputs can be objectively verified by programs through straightforward checks (e.g., checking if a speech transcript begins with a keyword by comparing the first word with the keyword), and (2) \emph{near-verifiable} tasks, which cannot be fully verified by programs but are brought close to verifiable by strengthening task conditions, e.g., providing reference inputs, and constraining model judges with explicit rubrics. These conditions create significant \emph{generation-evaluation gaps}, where the generation step is challenging due to complex constraints, yet the evaluation step reduces to a simple and objective check (e.g., generating an image of a snowman without a carrot nose can be challenging due to spurious correlation \citep{ye2024spurious}, but verifying the absence of the carrot nose can be achieved accurately by prompting a VLM with a targeted question). Since near-verifiable tasks are not strictly verifiable, we do not take their evaluation for granted but validate it against human judges. Example tasks can be found in Figure \ref{fig:example}.

\textsc{MMMG} includes 55 tasks (31 are newly developed) and 1288 instructions across 4 modality combinations---text, image, audio, and interleaved modalities---as depicted in Table \ref{tab:detail}. By categorizing tasks based on assessed capabilities, \textsc{MMMG} enables fine-grained analysis of model performance and targeted identification of weaknesses.

To validate the human alignment of \textsc{MMMG}, we conduct human evaluation across 41 tasks---978 instructions and 2596 evaluation questions---with each question assessed by three independent annotators. \textsc{MMMG} achieves an average human agreement of 94.4\% with average inter-annotator agreement being 97.1\%. Modality-specific agreements achieve 93.5\% for image, 92.3\% for audio, 96.6\% for interleaved image-text, and 91.0\% for interleaved audio-text, with relative improvements over prior best results by 12.7\% for image, and 34.5\% for interleaved image-text evaluation \citep{ghosh2023geneval, chen2024interleaved}.

We benchmark 29 open and proprietary multimodal generation models using the most human-aligned evaluation methods. Partial results are shown in Figure \ref{fig:bench_result}; the rest are in Appendix \ref{sec:full_res}. We find that modality-unified autoregressive models (ARMs) surpass diffusion models in image generation, with \textsc{GPT Image} \citep{openai2024gptimage} achieving the best accuracy of 70.7\%. However, \textsc{GPT Image} still falls short in interleaved text-image reasoning tasks, achieving only 10.1\% accuracy for math and code and 31.8\% for 3D scene transformation. Our qualitative analysis reveals that \textsc{Gemini 2 Image}, tends to tangle multiple images in generation, hindering accurate image-sequence and image-text pair generation. Additionally, \textsc{MMMG} reveals greater improvement potentials for audio generation compared to image, with top-performing models achieving accuracies of 48.7\% for sound and 46.3\% for music generation. Overall, to the best of our knowledge, \textsc{MMMG} is the \emph{first} multimodal generation benchmark built upon verifiable and near-verifiable tasks, and it provides the most comprehensive model ranking and fine-grained capability analysis while ensuring reliability.
\begin{table}[!t]\centering
    \scriptsize
    \tabcolsep=0.08cm
    \begin{tabular}{lcccccccccccccc}\toprule[1.5pt]
     \multirow{3}{*}{\textbf{Dataset}} &\multirow{3}{*}{\textbf{\# Samples}} &\multirow{3}{*}{\textbf{\# Tasks}} &\multicolumn{4}{c}{\textbf{Generation Modality}} &\multicolumn{4}{c}{\textbf{Evaluation}} &\multicolumn{3}{c}{\textbf{Tested Capability}} \\
     \cmidrule(lr){4-7} \cmidrule(lr){8-11} \cmidrule(lr){12-14}
    & & &\faImage &\faVolumeDown &$\mathbb{T}$ + \faImage &$\mathbb{T}$ + \faVolumeDown &human &mllm &score &code &gen &edit &reason \\\midrule[0.75pt]
    GenEval \citep{ghosh2023geneval} &553 &6 &\greentick &\redcross &\redcross &\redcross &\redcross &\redcross &\greentick &\greentick &\greentick &\redcross &\redcross \\
    DrawBench \citep{saharia2022photorealistic} &200 &11 &\greentick &\redcross &\redcross &\redcross &\greentick &\redcross &\redcross &\redcross &\greentick &\redcross &\redcross \\
    GenAI-Bench \citep{li2024genai} &1,600 &8 &\greentick &\redcross &\redcross &\redcross &\greentick &\redcross &\redcross &\redcross &\greentick &\redcross &\redcross \\
    AudioTime \citep{xie2024audiotime} &500 &4 &\redcross &\greentick &\redcross &\redcross &\redcross &\redcross &\question &\greentick &\greentick &\redcross &\redcross \\
    MusicEval \citep{liu2025musiceval} &384 &1 &\redcross &\greentick &\redcross &\redcross &\greentick &\redcross &\redcross &\redcross &\greentick &\redcross &\redcross \\
    CommonVoice \citep{ardila2019common} &58,250 &1 &\redcross &\greentick &\redcross &\redcross &\redcross &\redcross &\greentick &\redcross &\greentick &\redcross &\redcross \\
    $\text{MMIE}_{\text{MMG}}$ \citep{xia2024mmie} &16,487 &7 &\redcross &\redcross &\greentick &\redcross &\redcross &\question &\redcross &\redcross &\greentick &\redcross &\redcross \\
    CoMM \citep{chen2024comm} &227,000 &4 &\redcross &\redcross &\greentick &\redcross &\redcross &\question &\greentick &\redcross &\greentick &\redcross &\redcross \\
    ISG-Bench \citep{chen2024interleaved} &1,150 &21 &\redcross &\redcross &\greentick &\redcross &\redcross &\question &\greentick &\redcross &\greentick &\greentick &\redcross \\
    $\text{MixEval-X}_{\text{MMG}}$ \citep{ni2024mixeval} &600 &3 &\greentick &\greentick &\redcross &\redcross &\greentick &\question &\redcross &\redcross &\greentick &\greentick &\redcross \\
    Eval-Anything \citep{ji2024align} &500 &6 &\greentick &\greentick &\greentick &\redcross &\greentick &\question &\redcross &\redcross &\greentick &\redcross &\redcross \\
    \textsc{MMMG} (Ours) &1248 &55 &\greentick &\greentick &\greentick &\greentick &\redcross &\greentick &\greentick &\greentick &\greentick &\greentick &\greentick \\
    \bottomrule[1.5pt]
    \end{tabular}
\vspace{5pt}
\caption{MMMG compared with other multimodal generation benchmarks. \faImage, \faVolumeDown, $\mathbb{T}$ + \faImage, $\mathbb{T}$ + \faVolumeDown \ represent image, audio, interleaved image-text, and interleaved audio-text generation, respectively. ``score'' stands for embedding-based / rule-based similarity score, ``code'' for programmatically verification, and ``reason'' for multi-step reasoning. \question \ represents low human alignment or no human experiments. \textsc{MMMG} exceeds other benchmarks in the number of covered tasks and modalities while providing more reliable evaluation. } \label{tab:compare}
\vspace{-10pt}
\end{table}

\section{Related Work}
\textbf{Interleaved Multimodal Generation.} Interleaved multimodal generation involves generating coherent content across multiple modalities simultaneously, such as visual storytelling \citep{huang2016visual, wen2023infovisdial}, reference-based image editing \citep{chen2025multimodal}, and voice chatbots \citep{chu2024qwen2}. Effective models must understand multimodal inputs and produce aligned outputs across modalities. Current approaches include (1) LLM backbones with specialized decoders \citep{chen2025janus, xie2024show}, which leverage dedicated components to render visual or audio outputs; (2) modality-unified autoregressive models \citep{chern2024anole, hurst2024gpt, wang2024emu3}, processing text, visual, and acoustic tokens within a single sequence model, enabling native generation of interleaved content; and (3) agent-based methods \citep{chen2024interleaved}, using a ``Plan-Execute-Refine'' pipeline with modality-specific tools. Despite significant advances, evaluation frameworks for interleaved multimodal generation remain underdeveloped, particularly in accurately and automatically assessing cross-modal consistency, and instruction-following capabilities.

\textbf{Multimodal Generation Evaluation.} Evaluating image, audio and their interleaved generation presents unique challenges that have been addressed through several approaches, each with notable limitations, including (1) using specialized visual or audio models \citep{ghosh2023geneval, xie2025audiotime}, which struggle to generalize beyond their training data \citep{ming2022delving}; (2) directly employing MLMs as evaluators \citep{xia2024mmie, chen2024comm, chen2024interleaved}, which often misalign with human judgments \citep{chen2024mllm}; and (3) for image evaluation particularly, leveraging visual question answering (VQA) to assess specific aspects of generated content \citep{hu2023tifa, lin2024evaluating}, which declines significantly in accuracy when facing complex evaluation scenarios that require nuanced reasoning \citep{chen2024interleaved}. To address these limitations, previous research incorporates extensive human preference data to enhance MLM accuracy \citep{xiong2024llava, yao2024varying}. Our work is an orthogonal approach that carefully designs evaluation instructions to leverage current MLM strengths while mitigating their limitations, enabling reliable multimodal evaluation without extra training or finetuning. Table \ref{tab:compare} compares \textsc{MMMG} with existing benchmarks.

\textbf{Verifiable Instructions.} In text generation, verifiable instructions, i.e., constraints whose satisfaction can be objectively checked by programs such as response length or keyword inclusion, enable reproducible evaluation without human or model judges \citep{zhou2023instruction}. Such verifiable signals have further become central to LLM post-training, e.g., reinforcement learning with verifiable rewards on instruction following, math, and code \citep{lambert2024tulu, guo2025deepseek}. In contrast, multimodal generation has lacked verifiable benchmarks, as the correctness of an image or audio clip rarely reduces to a simple rule. \textsc{MMMG} takes a first step in bringing this paradigm to multimodal generation: it includes programmatically verifiable tasks where possible, and complements them with near-verifiable tasks whose strengthened conditions make model-based evaluation objective and highly human-aligned.

\section{MMMG Benchmark Construction}
\label{sec:mmmg}

Our goal is to build a multimodal generation benchmark that (1) covers a wide range of modalities and their combinations (image, audio, interleaved text and image, interleaved text and audio) with diverse tasks spanning different model capabilities. For each task, (2) we also ensure it is verifiable or near-verifiable, so that automated evaluation is reliable and aligns well with humans. In this section, we first discuss our data and instruction construction in detail (\S\ref{sec:mmmg:data}), and then introduce the evaluation methods we built for each task (\S\ref{sec:mmmg:metrics}).

\subsection{Data Curation}
\label{sec:mmmg:data}
To guarantee high-quality instructions and reliable evaluation, we design a systematic data curation pipeline consisting of three key stages.

\textbf{Task Creation.} We begin by creating an initial pool of 78 candidate task templates. These tasks span various modality combinations and each task aims to evaluate a single multimodal generation capability. The complete list of 76 tasks can be found in Appendix \ref{sec:all_tasks}. We verify if each task is testing different model capability in Appendix \ref{sec:analysis}. For each task, we conduct a rigorous feasibility assessment to ensure there is at least one reliable evaluation method---either programmatic verification (verifiable) or a highly human-aligned evaluation method (near-verifiable). To make tasks near-verifiable, we strengthen their conditions so that each task is formulated with a single ground-truth answer and can be evaluated objectively. Based on this process, we narrow our task pool down to 60 tasks.

\textbf{Instruction Synthesis and Validation.} We hire experts to write templates and 5 seed instructions per task. Then we employ a human-in-the-loop approach to synthesize high-quality instructions for each task. Inspired by Self-Instruct \citep{wang2022self}, we prompt \textsc{GPT-4o} \citep{hurst2024gpt} with the task template and quality-controlled criteria to generate 10 candidate instructions per task. We then go through a two-stage selection process:

\begin{itemize}[leftmargin=*]
    \item \textbf{Quality Filtering.} Initially, we remove instructions that are ambiguous (instructions with unclear or multiple interpretations), unrealistic (instructions that describe improbable or nonsensical scenarios), or redundant (instructions that closely resemble previously accepted examples). For instance, unrealistic instruction ``\emph{Generate an image of a forest without any trees}'' is discarded because it is semantically contradictory and unlikely to occur in actual user queries.
    \item \textbf{Verifiability Assessment.} After quality filtering, we sample generated outputs and verify if auto evaluation aligns with human judgments to avoid cases where models fail to accurately evaluate out-of-distribution samples. For example, \textsc{GPT-4o} can accurately count fewer than 10 objects but is prone to errors counting more than 10.
\end{itemize}

We then generate another 10 candidate instructions and repeat the generation and validation process continues until we gather approximately 40 high-quality instructions per task. Statistically, only 10\% of generated instructions pass examination, highlighting the high standard of data selection.

\textbf{Postprocessing.} For final quality control, we perform a task-level postprocessing step to further refine our benchmark. This involves two procedures: (1) Task filtering: we recruit two independent annotators to judge if each task is realistic. We eliminate six tasks that at least one annotator judges to be unrealistic. (2) Instruction paraphrasing: To ensure linguistic diversity and prevent models from memorizing specific instruction patterns, we paraphrase all remaining instructions. Each paraphrased instruction is examined manually to verify that it is equivalent to the original instruction semantically.

To this end, we ultimately collect a total of 1288 instructions across 55 tasks spanning 4 modality combinations. This systematic approach ensures that \textsc{MMMG} provides a comprehensive, fine-grained, and reliable evaluation framework for assessing multimodal generation capabilities. The detailed metadata of each task in MMMG is in Table \ref{tab:detail}.

\subsection{Evaluation Method}
\label{sec:mmmg:metrics}

We report the evaluation method used for each task in Table \ref{tab:detail}. Broadly, tasks checked purely by programs are verifiable, while tasks that involve model-based evaluators (VLMs, audio models, or embedding similarity) are near-verifiable; for the latter, the designs below further tighten the evaluation conditions to approach the reliability of programmatic checks.

\textbf{VLM.} We employ vision language models (VLMs) for most reference-free image evaluation tasks as shown in Figure \ref{fig:example}(b), since they generalize better than object detection or OCR models in out-of-domain scenarios. Instead of letting a VLM freely judge a generation, we use \textbf{format rubrics} and \textbf{content rubrics} that specify both \emph{how} it should answer and \emph{what} it should verify. Specifically, the format rubrics constrain the response format. For example, in boolean judgments, chain-of-thought prompting, which explicitly instructs the model not to output yes/no at the beginning of its response, significantly improves judgment reliability. For counting and spatial reasoning, we find multiple-choice questions much more effective. We hypothesize that multiple-choice questions reduce the output space and thereby simplify these tasks. For example, including an option like ``\emph{E. More than 6}'' in object counting questions can prevent miscounting errors when many objects appear in the image. The content rubrics specify required objects, attributes, and relations as positive rubrics, and also include explicit negative conditions, using concrete rather than abstract wording. For example, unlike only asking for ``\emph{one basketball with a cube shape}'', ``\emph{one basketball with a cube shape instead of a sphere}'' makes the rejection criterion much clearer. Compared with automatically generated VQA pairs such as TIFA \citep{hu2023tifa}, human-designed rubrics reduce hallucinations, format errors, and near-miss acceptances, enabling more reliable VLM-as-a-judge, especially on challenging tasks.

\textbf{Image Similarity.} For reference-based image evaluation tasks requiring perceptual similarity, we employ DreamSim \citep{fu2023dreamsim}. When exact matching is necessary, we use SSIM \citep{wang2004image}. For image editing tasks, we implement a dual approach: DreamSim/VQA evaluates the edited region, while SSIM assesses the unmodified areas outside it, ensuring that local editing instructions are precisely followed as shown in Figure \ref{fig:example}(e).

\textbf{Audio Similarity.} Research indicates that current audio language models (ALMs) cannot reliably analyze sound or music clips \citep{sakshi2024mmau}. Therefore, we select ESC-50 \citep{piczak2015esc}, OpenMIC-2018 \citep{humphrey2018openmic} and GTZAN \citep{sturm2012analysis} as reference datasets for sound and music evaluation, and compute the average top-10 CLAP cosine similarity \citep{wu2023large} with reference audio as shown in Figure \ref{fig:example}(c).

\textbf{Audio Model.} For specialized audio analysis, we employ several targeted models. \textsc{WavLM} \citep{chen2022wavlm} is employed for speaker similarity verification. For speech transcription, we use Whisper \citep{radford2023robust} as shown in Figure \ref{fig:example}(d). Gender classification in speech leverages a finetuned \textsc{Wav2Vec} checkpoint \citep{fiury2023wav2vec2gender}. For music tempo computation, we employ \textsc{BeatThis} \citep{foscarin2024beat} for beat tracking and the beats statistics are used for music tempo computation.

\textbf{Program.} For programmatic verification, we utilize PIL for image analysis as shown in Figure \ref{fig:example}(a), Librosa \citep{mcfee2015librosa} and Praat \citep{boersma2001speak} for audio pitch, intensity, and speed analysis. For textual constraint verification, we follow the implementation of IF-Eval \citep{zhou2023instruction}. We use word accuracy (WAcc) to evaluate textual similarity for text rendering and text-to-speech tasks which requires exact matching.

\textbf{Scoring.} Each generation receives either a binary classification or an accuracy score reflecting instruction-following level. Binary labels are converted to numerical scores (0.0 for incorrect, 1.0 for correct), and all task scores are macro-averaged following prior work \citep{ghosh2023geneval}. Despite employing different evaluation protocols, \textsc{MMMG} guarantees implicit unification across protocols by high human alignment. For example, even though CLAPScore gives scores between [0, 1], task-specific thresholds are selected to best align with human judgment. This makes CLAPScore's sensitivity equivalent to VLM binary judgments because both are calibrated against the same unified human judgment standard.

% \section{Experiment Settings}
\begin{figure}[!t]
    \centering
    \begin{subfigure}[b]{0.48\textwidth}
        \centering
        \includegraphics[width=\textwidth]{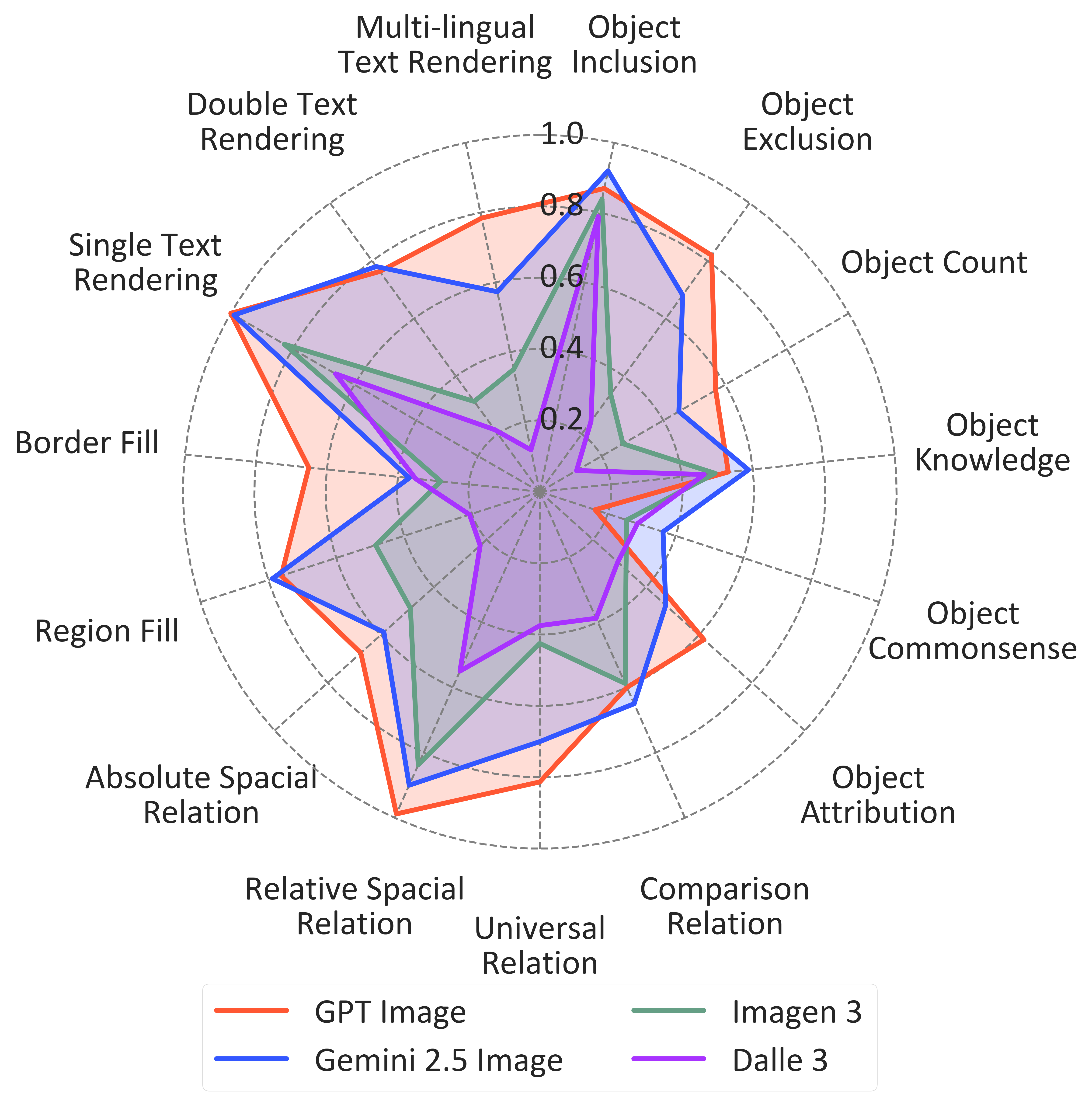}
        \caption{Image Generation}
    \end{subfigure}
    \hfill
    \begin{subfigure}[b]{0.48\textwidth}
        \centering
        \includegraphics[width=\textwidth]{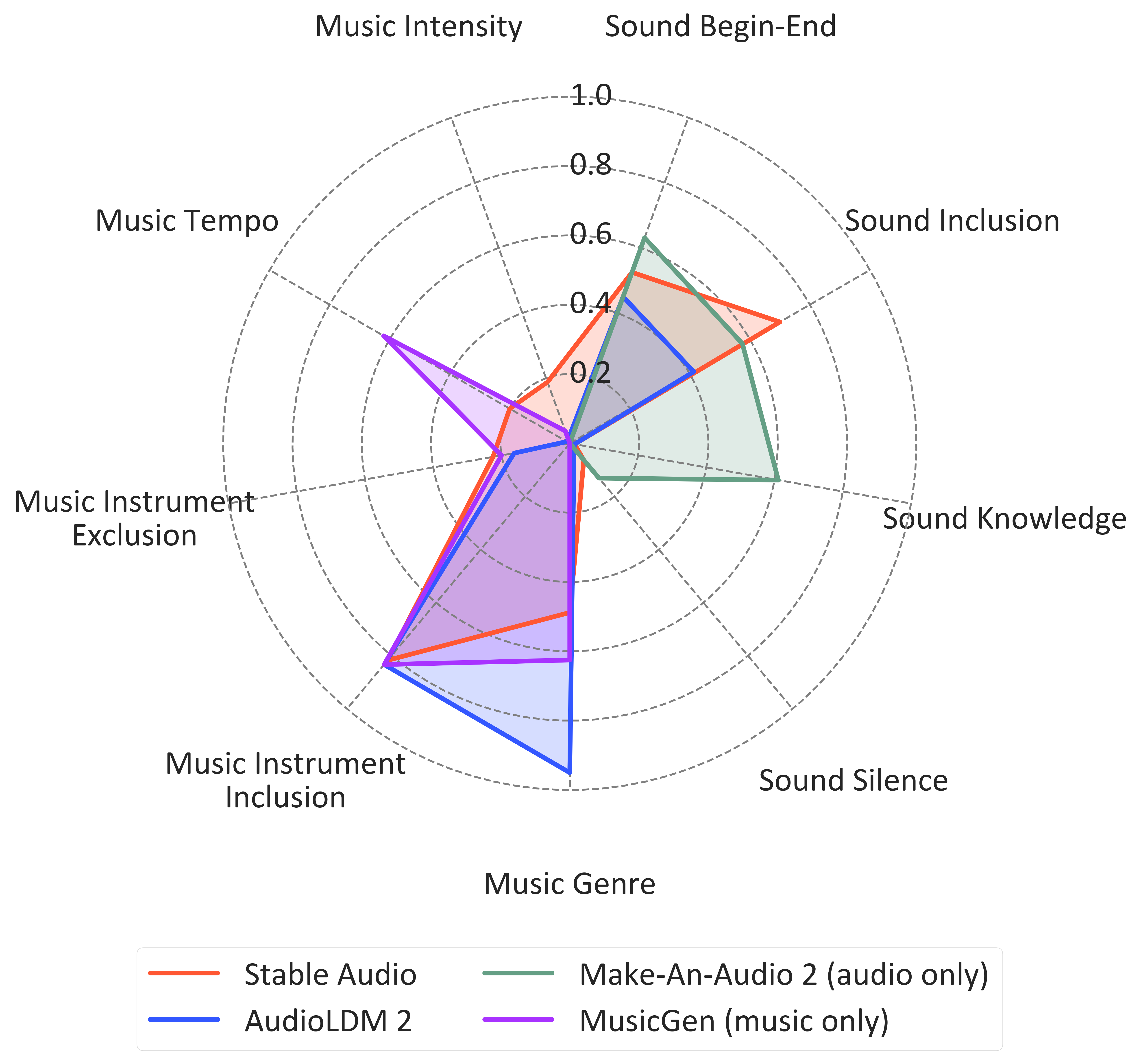}
        \caption{Sound and Music Generation}
    \end{subfigure}
    
    \begin{subfigure}[b]{0.48\textwidth}
        \centering
        \includegraphics[width=\textwidth]{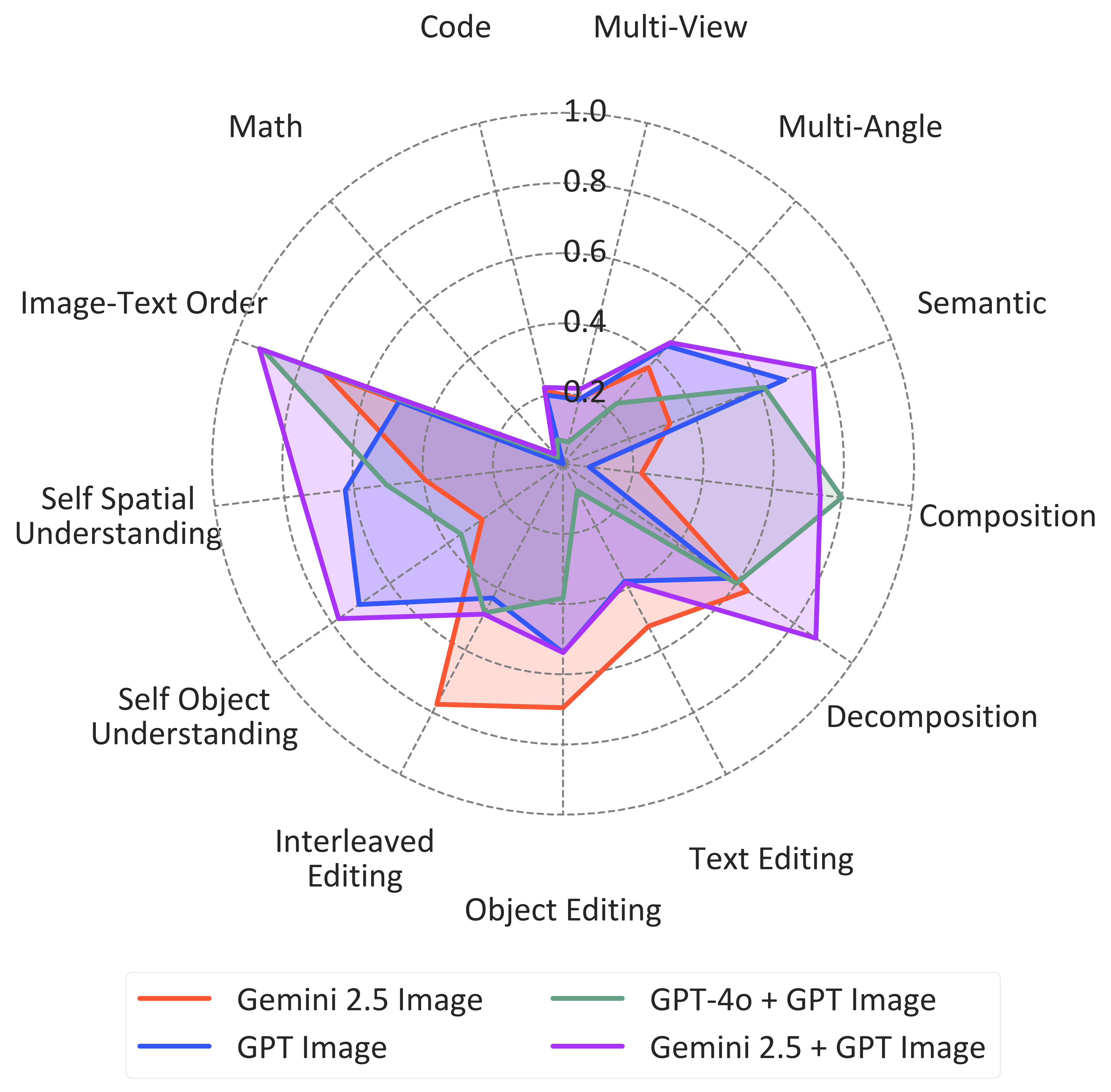}
        \caption{Interleaved Image-Text Generation}
    \end{subfigure}
    \hfill
    \begin{subfigure}[b]{0.48\textwidth}
        \centering
        \includegraphics[width=\textwidth]{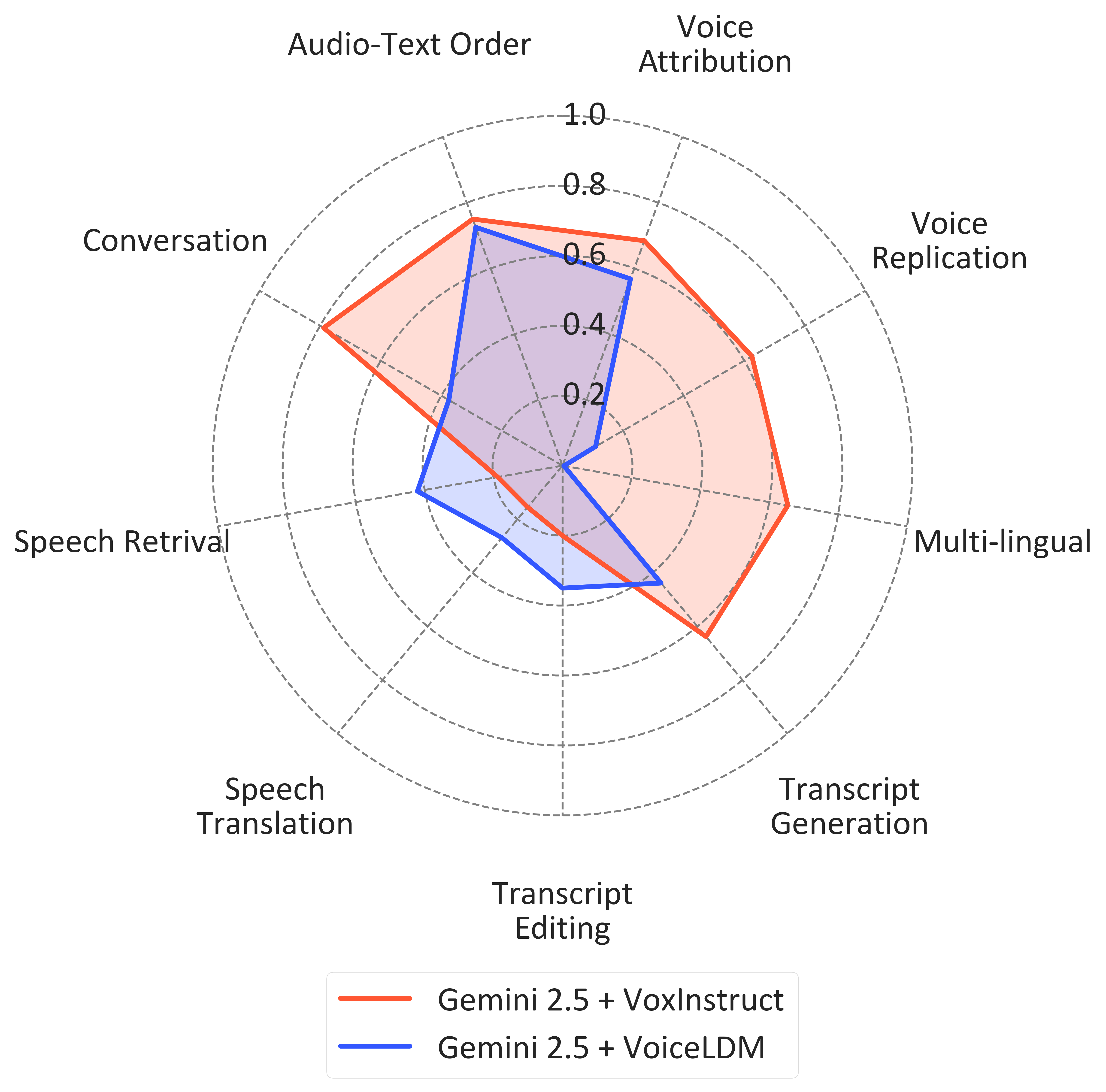}
        \caption{Speech and Interleaved Speech-Text Generation}
    \end{subfigure}
    
    \caption{Benchmark results of multimodal generation models on \textsc{MMMG} covering four modality combinations. Please refer to Table \ref{tab:detail} for more detailed category information. We aggregate some sub-tasks for interleaved image-text generation. \textsc{GPT Image} beats all other models on most image generation tasks, and strongly competes other baselines in generating consistent image sequences and coherent interleaved image-text contents.}
    \label{fig:bench_result}
    \vspace{-10pt}
\end{figure}

\section{Results and Analysis}
In this section, we report human annotation results in \S \ref{sec:exp_human}, benchmarking results evaluated by the most human-aligned metrics in \S \ref{sec:exp_benchmark}. We also report the correlation with real-world human preference leaderboard in \S \ref{sec:exp_leaderboard}. For detailed experiment setup, including model, prompt, evaluation and annotation implementation, please refer to Appendix \ref{sec:detail_exp}.

\subsection{Alignment with Human Judges}
\label{sec:exp_human}
We conduct human evaluations on 978 instructions evaluated by models. For each instruction, we randomly select two models from all models evaluated on this instruction and obtain one generation per model. Each generation is evaluated by two independent annotators, randomly selected from our pool of 20 graduate student annotators. To reduce subjective bias, we design specific multiple-choice questions for each instruction exemplified in Appendix \ref{sec:human_question}, thereby constraining annotators' responses to a fixed set of choices and ensuring high inter-annotator agreement. In cases of disagreement, a third annotator determines the final annotation. In total, human studies involve 2596 evaluation questions and collect 5192 annotations. For strictly verifiable instructions, human alignment validation is unnecessary as they are checked programmatically; we therefore focus human validation on near-verifiable instructions, whose model-based evaluation is not guaranteed to be correct. Human-model and inter-human agreement measures can be found in Table \ref{tab:combined}. 

\textsc{MMMG} demonstrates high human alignment, with average best human-model agreement for image, audio, interleaved image-text, and interleaved audio-text being 0.935, 0.923, 0.966 and 0.910 respectively, calculated by selecting the most human-aligned evaluation method averaging across tasks. The average inter-annotator agreement is as high as 0.971, with a worst-case value of 0.917. This proves the objectivity of \textsc{MMMG} and alleviates the subjective annotation concern. Together, these results suggest that strengthened task conditions allow model-based evaluation on near-verifiable tasks to be highly reliable, though not error-free. \textsc{MMMG} also outperforms previous best benchmark alignment significantly: agreement on image generation surpasses GenEval (0.830) by 12.7\%, and Pearson correlation on interleaved image-text generation surpasses ISG-Bench (0.718) by 34.5\%. Experiments show that \textsc{GPT-4o} remains the most human-aligned image evaluation model with an average agreement of 0.935. Though open-source models like \textsc{Qwen2.5-VL} still have a gap with proprietary models, its judging accuracy already surpassing previous SOTA benchmarks, suggesting the high reliability of \textsc{MMMG}. For audio evaluation, even though $\mathrm{CLAPScore}_{\mathrm{text}}$ yields a satisfactory agreement of 0.923, it relies highly on the quality of reference audio, thus making it challenging for out-of-domain audio evaluation. 

% TODO
\subsection{Benchmarking Results}
\label{sec:exp_benchmark}

Selected model performances are illustrated in Figure \ref{fig:bench_result}, with complete evaluation results provided in Appendix \ref{sec:full_res}.

\textbf{Image Generation.} ARMs outperform diffusion models significantly, with \textsc{GPT Image} and \textsc{Gemini 2.5 Image} achieving accuracies of 0.707 and 0.654 respectively, ranking 1st and 2nd. This indicates that ARMs with stronger linguistic capabilities can better follow instructions. However, models struggle notably when generating objects with uncommon attributes or unusual relationships, showing average accuracies of only 0.340 and 0.416 respectively. This underscores the vulnerability of image generation models to out-of-domain instructions. Despite top models (\textsc{GPT Image}) achieving 0.531 accuracy on knowledge reasoning tasks, they fail drastically on commonsense reasoning (0.163 accuracy), suggesting their performance may rely more on memorization than genuine reasoning ability.

\textbf{Interleaved Image-Text Generation.} Ensembling \textsc{Gemini 2.5} improves \textsc{GPT Image}'s accuracy by 21.1\%, indicating that agent-based models outperform unified ARMs via stronger planning capabilities and clear modality separation, which leads to better image consistency and coherence. Still, some tasks pose considerable challenges, with the best-performing combination (\textsc{Gemini 2.5} + \textsc{GPT Image}) achieving limited accuracies of 0.131 on math and coding reasoning, 0.341 on 3D scene transformations, and 0.383 on text editing. Error analysis on one of the ARM, \textsc{Gemini 2 Image}, reveals that it suffers from two primary failure modes: (1) misinterpreting image order in interleaved inputs, and (2) tangle multiple images in output due to continuous latent representations, as shown in Figure \ref{fig:failure}. However, \textsc{Gemini 2.5 Image} as an ARM perform best at image editing tasks with an accuracy of 0.672, which may indicate that ARMs preserve input image information better due to reduced information loss from image-to-text transitions.

\textbf{Sound and Music Generation.} Only \textsc{Make-An-Audio 2}, which leverages LLMs for instruction parsing, shows competence in sound reasoning. Other models exhibit reasoning limitations, with accuracies of 0.233 for instrument exclusion and 0.175 for souQnd knowledge reasoning. Volume control is also poor, with silence generation and intensity control reaching 0.048 and 0.063 accuracy, respectively. Only \textsc{MusicGen} effectively handles tempo control, while only \textsc{Stable Audio} and \textsc{AudioLDM 2} support both sound and music generation, suggesting that audio generation models remain largely domain-constrained.

\textbf{Speech and Interleaved Speech-Text Generation.} \textsc{Spirit LM}, the only inherently interleaved speech-text model, fails entirely on most speech generation tasks. Agent-based models struggle with tasks requiring simultaneous speech understanding and generation, reaching average accuracies of 0.275 for speech editing, 0.212 for speech translation, and 0.304 for speech retrieval. Ablation studies reveal that LLM backbones perform perfectly on speech transcripts, indicating that failures stem from error accumulation in speech processing rather than reasoning deficits.

\subsection{Correlation with Real-World Leaderboard}
\label{sec:exp_leaderboard}

We compare the correlation of the \textsc{MMMG} score with the Chatbot Arena \citep{chiang2024chatbot} score on the text-to-image task. We take the Arena Score for 9 image generation models under the ``User Prompts Only'' category as a gold reference. We report the Pearson correlation and Spearman's rank correlation coefficient between gold arena scores and scores produced by evaluating on different benchmarks in Table \ref{tab:chatbot}. We compare with GenEval, DrawBench, and GenAI-Bench. We employ VQAScore \citep{lin2024evaluating} to replace human evaluation on DrawBench and GenAI-Bench; due to budgetary limitations, we randomly sample 400 out of 1600 instructions for GenAI-Bench. 

\begin{wraptable}{r}{0.5\linewidth}{
\vspace{-10pt}
\centering
\resizebox{1\linewidth}{!}{
\scriptsize
\tabcolsep=0.15cm
\begin{tabular}{lccccc}\toprule[1.5pt]
\textbf{Model} &\textbf{Arena} &\textbf{GenEval} &\textbf{Draw}
&\textbf{GenAI} &\textbf{MMMG} \\\midrule[0.75pt]
\textsc{Imagen 3} &1064 &0.707 &0.861 &0.793 &0.474 \\
\textsc{Recraft v3} &1018 &0.732 &0.826 &0.817 &0.441 \\
\textsc{Luma Photon} &997 &0.738 &0.766 &0.804 &0.587 \\
\textsc{Flux 1.1 Pro} &992 &0.588 &0.725 &0.736 &0.431 \\
\textsc{Ideogram 2} &1011 &0.615 &0.757 &0.782 &0.508 \\
\textsc{Dalle 3} &978 &0.627 &0.809 &0.811 &0.352 \\
\textsc{SD 3.5} &911 &0.591 &0.711 &0.715 &0.332 \\
\textsc{Gemini 2 Image} &996 &0.669 &0.765 &0.783 &0.592 \\
\textsc{GPT Image} &1126 &0.808 &0.793 &0.824 &0.707 \\
Spearman & &0.460 &0.444 &0.418 &\textbf{0.561} \\
\bottomrule[1.5pt]
\end{tabular}}
\caption{Correlation of automated image generation benchmarks with Chatbot Arena. Arena, Draw, GenAI represent Chatbot Arena, DrawBench, and GenAI-Bench. MMMG achieves the highest correlation with Chatbot Arena. This indicates even though our instructions are synthetic, the evaluation results are still highly human-aligned.}
\label{tab:chatbot}
\vspace{-10pt}
}
\end{wraptable}

\textsc{MMMG} provides reliable model rankings with a Spearman correlation coefficient of 0.561, significantly outperforming baseline benchmarks. This indicates that despite that synthetic instructions may not fully align with real-world queries, \textsc{MMMG} achieves higher alignment with human preferences. Such results suggest that evaluator alignment (i.e., the reliability of the evaluation method) may outweigh instruction distribution alignment (i.e., the extent to which benchmark tasks reflect real-world task distributions) for accurate model assessment. Moreover, \textsc{MMMG} demonstrates superior differentiation capabilities among evaluated models. The performance gap of 0.375 between the highest- and lowest-ranked models is much larger than the next-best baseline (GenEval), which has only a gap of 0.217. This larger range underscores \textsc{MMMG}'s enhanced ability to distinguish among models, particularly for differentiating performance among top-tier models.

Due to the lack of real-world human preference leaderboards like Chatbot Arena for other modalities, we leave correlation studies for other modalities as future work.

\section{Conclusion}
In this work, we introduce \textsc{MMMG}, a comprehensive automated evaluation suite for multitask multimodal generation and, to the best of our knowledge, the first to bring verifiable tasks to multimodal generation, complemented by near-verifiable tasks whose strengthened conditions enable reliable evaluation. We collect 1288 high-quality instructions spanning 55 diverse tasks involving text, image, audio, and interleaved content. Extensive human validation demonstrates that \textsc{MMMG} correlates better with human judgments compared to previous benchmarks. Benchmarking results highlight ongoing challenges in multimodal reasoning, interleaved generation, and audio generation. The fine-grained nature of \textsc{MMMG} enables detailed capability analysis, providing valuable insights for targeted multimodal improvements. Beyond serving as a leaderboard, we hope the verifiable and near-verifiable design of \textsc{MMMG} inspires scalable collection of reliable reward signals for future multimodal generation training.
%Given the page limit, we refer readers to Appendix \ref{sec:limitation} for limitations and social impacts discussion.

\newpage
\bibliography{colm2026_conference}
\bibliographystyle{colm2026_conference}

\newpage
\appendix

\begin{figure}[!t]
    \centering
    \includegraphics[width=1\linewidth]{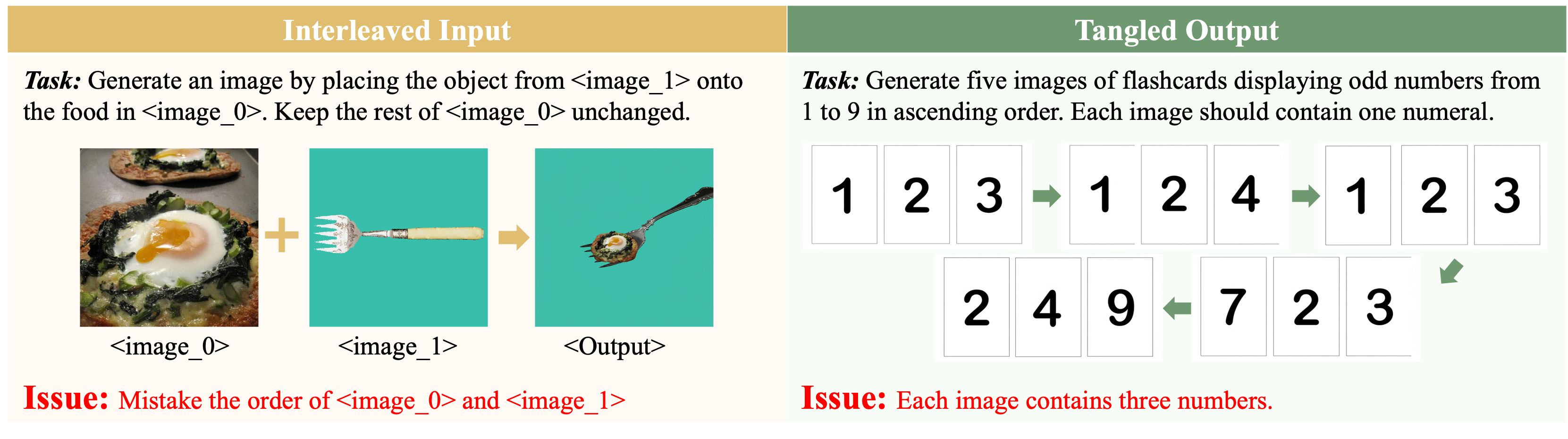}
    \caption{Two prevalent failure cases observed in interleaved image-text generation tasks for \textsc{Gemini 2 Image}: (1) models fail to accurately interpret the order of images in interleaved inputs; and (2) models frequently blend multiple images together, possibly due to limitations in encoding multiple images with continuous latent image representations.}
    \label{fig:failure}
\vspace{-10pt}
\end{figure}

\section{Limitations and Ethics Statement}
\label{sec:limitation}
\subsection{Limitations}
While \textsc{MMMG} constitutes a significant advancement in automated multimodal generation evaluation, we acknowledge several limitations inherent to our methodology and scope.
\paragraph{Limited Task Coverage.} \textsc{MMMG} does not exhaustively cover all potential tasks within multimodal generation, particularly in the domains of interleaved image-text generation and sound/music generation. This limitation primarily arises from current inadequacies in available evaluation methods or models, which fail to yield sufficiently human-aligned results on numerous widely-used tasks. Such gaps in coverage may introduce biases into our model rankings, potentially misaligning evaluation results with actual user experiences. To mitigate this, we intend to dynamically expand and update our benchmark tasks in real-time as more powerful and reliable evaluation models become available. We also include tasks that we considered commonly used but abandoned due to infeasible evaluation in Appendix \ref{sec:all_tasks}.
\paragraph{Dependence on Proprietary Models.} Our evaluation relies on proprietary models (e.g., \textsc{GPT-4o}, \textsc{Gemini 2.5}). The substantial performance gap between proprietary and open-source models makes reliance on proprietary models necessary for achieving highly accurate and human-aligned evaluations across diverse tasks. Unfortunately, current open-source alternatives often lack sufficient accuracy on certain complex tasks, rendering them unsuitable as reliable evaluators. Consequently, this dependence limits broad reproducibility and access within the academic community, highlighting the urgent need for improved and accessible open-source evaluation models.

\subsection{Ethics Statement}  \textsc{MMMG} is designed as a general-purpose benchmark for evaluating multimodal generation capabilities and deliberately excludes high-stakes applications, e.g. medical image generation or other safety-critical domains that would require specialized evaluation protocols, domain expertise, and more rigorous validation procedures. For researchers and practitioners deploying multimodal generation models in high-stakes scenarios, we emphasize that \textsc{MMMG}'s evaluation framework should not be considered sufficient (even with reported SOTA human agreement on general domains) without additional domain-specific validation, expert review, and safety protocols appropriate to the specific application context.

\subsection{Reproducibility Statement}
We provide comprehensive details to ensure full reproducibility of MMMG's benchmark construction and evaluation. Appendix \ref{sec:source} documents all data sources with specific dataset versions and access links. Appendix \ref{sec:detail_model} specifies all 29 evaluated generation models with exact checkpoint identifiers, along with 3 VLMs and 4 specialized audio models used for evaluation. Detailed evaluation protocols are provided in Appendix \ref{sec:detail_eval}, including task-specific VLM prompts, programmatic verification procedures using PIL, Librosa, and Praat, etc., and agent-based generation system prompts (Tables \ref{tab:i_agent_prompt} and \ref{tab:a_agent_prompt}). Human annotation procedures are documented in Appendix \ref{sec:human_question} with example annotation interfaces (Figure \ref{fig:interface}) and task-specific annotation questions. Computational requirements and costs are detailed in Appendix \ref{sec:compute}, including GPU specifications, runtime estimates and API costs. All benchmark data, evaluation code, and model outputs are publicly available at \href{https://anonymous.4open.science/r/MMMG-3530/}{this anonymous link}. We append instruction and model generation examples in the supplementary material.

\subsection{Use of Large Language Models (LLMs) Statement} For paper writing, we used LLMs (specifically \textsc{CLaude 4}) solely for language polishing and improving clarity after drafting the complete manuscript ourselves. All factual claims, experimental results, analyses, and conclusions were written by the authors and carefully verified for accuracy before and after any LLM-assisted editing. For data collection, we employed \textsc{GPT-4o} to synthesize part of the candidate instructions as described in Section \ref{sec:source}. These LLM-generated instructions underwent rigorous quality filtering and verifiability assessment by human annotators as stated in Section \ref{sec:mmmg:data}. Additionally, we occasionally used LLMs to assist with debugging experimental code.

\section{Detailed Dataset Information}
\subsection{Data Source}
\label{sec:source}
\begin{itemize}[leftmargin=*]
    \item \textbf{Object Reasoning.} We sample from HotpotQA \citep{yang2018hotpotqa} through \href{https://hotpotqa.github.io/}{the official website} and WISE \cite{niu2025wise} through \href{https://github.com/PKU-YuanGroup/WISE}{the official website}. We take the QA pairs where the answers are individual objects and can be directly transformed into image generation instructions or the answers are nations and can be transformed into the national flags or animals generation instructions.
    \item \textbf{Image Editing.} We sample images from EmuEdit \citep{sheynin2024emu} through the ``facebook/emu\_edit\_test\_set'' checkpoint on Huggingface \citep{wolf2019huggingface} for object adding, removing, modifying, local, color and text editing tasks. We modify the instructions to make sure they are clear, unambiguous and more challenging. We also sample object images from COCO \citep{lin2014microsoft} through \href{https://cocodataset.org/#home}{the official website} and use PhotoShop to combine with the scene images in EmuEdit to form golden reference images. We sample scene images from CLEVR \citep{johnson2017clevr} through \href{https://cs.stanford.edu/people/jcjohns/clevr/}{the official website} for the interleaved color modifying task, since modifying color for pure-colored geometries is much more unambiguous than regular objects. We also use PhotoShop to generate the golden reference images.
    \item \textbf{3D transformation.} We sample instructions and golden reference images from ISG-Bench \citep{chen2024interleaved} through \href{https://github.com/Dongping-Chen/ISG}{the official website}. We polish the instructions to make sure they are clear and unambiguous.
    \item \textbf{Math.} We sample images from MM-IQ \citep{cai2025mm} through the ``huanqia/MM-IQ'' checkpoint on Huggingface. We manually edit the images to transform the multiple-choice questions into free-form generation questions. We have 2 annotators to check if the free-form questions can only have one possible answer without alternatives.
    \item \textbf{Code.} We sample SVG codes from StarVector \citep{rodriguez2023starvector} through the ``starvector/text2svg-stack'' checkpoint on Huggingface. and transform the original image-to-text instructions into interleaved reasoning instructions. We samples SVG codes with a length between 1000-1500 characters to control difficulty. We sample Tik (LaTeX) codes from DaTikZ \citep{belouadiautomatikz} through the ``HuggingFaceM4/datikz'' checkpoint on Huggingface.
    \item \textbf{Sound Generation.} We make sure all the target sounds fall in the 50 categories in ESC-50 \citep{piczak2015esc} through the ``ashraq/esc50'' checkpoint on Huggingface so that $\mathrm{CLAPScore}_{\mathrm{audio}}$ can have reference audios to compare with.
    \item \textbf{Instrument Generation.} We make sure all the target instruments fall in the 20 categories in OpenMIC-2018 \citep{humphrey2018openmic} through \href{https://github.com/cosmir/openmic-2018}{the official website} so that $\mathrm{CLAPScore}_{\mathrm{audio}}$ can have reference audios to compare with.
    \item \textbf{Speech.} For speech replication task, we samples speaker voices from LibriSpeech \citep{panayotov2015librispeech} ASR corpus through \href{https://www.openslr.org/12}{the official website} and use them as reference speeches for voice replication tasks. For speech translation task, we sample original speeches and human-annotated translation from CVSS dataset \citep{jia2022cvss}. For speech retrieval tasks, we sample from TED-LIUM 2 \citep{rousseau2014enhancing} for the original extremely long speech contexts.
\end{itemize}
The remaining tasks are generated from \textsc{GPT-4o} with manually designed templates.
\subsection{Excluded Tasks}
\label{sec:all_tasks}
We present the remaining 23 tasks we considered from our initial task set in Table \ref{tab:other_task}. We exclude ``Format Color'', ``Format Symmetric'', ``Speech Encoding'' tasks since they are not commonly seen in real user queries and ``Image-to-Sound'' and ``Sound-to-Image'' tasks are excluded because no models today can support these modalities. Other tasks are excluded because we could not find any reliable evaluation methods for those tasks.
\begin{table}[!ht]\centering
    \scriptsize
    \renewcommand*{\arraystretch}{1.2}
    \begin{tabular}{lp{8.5cm}cc}\toprule[1.5pt]
        \textbf{Task} & \textbf{Example} & \textbf{Input} & \textbf{Output}\\\midrule[0.75pt]
    Table Generation & Create a 2x2 table image. In the first column, place the text 'apple' in the top cell and 'pear' in the bottom cell. In the second column, place an image of an apple in the top cell and an image of a pear in the bottom cell.& $\mathbb{T}$ & \faImage \\
    Figure Generation & Create a histogram to visualize the given data. <data>& $\mathbb{T}$ & \faImage \\
    Format Color & Create a watermelon farm using only varying shades of red.& $\mathbb{T}$ & \faImage \\
    Format Symmetric & Generate an image of a futuristic cityscape. The image must be axisymmetric along the vertical center line.& $\mathbb{T}$ & \faImage \\
    Art Style & Create a painting of a dandelion sea in Impressionist style. & $\mathbb{T}$ & \faImage \\
    Photography & Create q zoomed out photo of a small bag of coffee beans from below. & $\mathbb{T}$ & \faImage \\
    Scene Editing & Make the weather in <image\_0> sunny. & $\mathbb{T}$, \faImage & \faImage \\
    Sound Count & Generate an audio of exactly three door knocks. & $\mathbb{T}$ & \faVolumeDown \\
    Sound Order & Generate an audio of a can being opened followed by a sipping sound. & $\mathbb{T}$ & \faVolumeDown \\
    Sound Duration & Generate audio of a car horn lasting for 3 seconds. & $\mathbb{T}$ & \faVolumeDown \\
    Speech Emotion  & Generate an audio of a woman sorrowfully saying, "What a life."  & $\mathbb{T}$ & \faVolumeDown \\
    Speech Accent  & Generate an audio of a man speaking in Indian accent, "What a beautiful day!"  & $\mathbb{T}$ & \faVolumeDown \\
    Speech Background  & Generate an audio of a man speaking in noisy train station distantly, "I am really busy."  & $\mathbb{T}$ & \faVolumeDown \\
    Speech Stress  & Generate an audio of a man saying, "Give me money now!" with stress on word "now".  & $\mathbb{T}$ & \faVolumeDown \\
    Music Emotion & Generate a vibrant, pulsating disco drum track. & $\mathbb{T}$ & \faVolumeDown \\
    Music Lyrics & Create a flute melody with the lyrics, <lyrics>. & $\mathbb{T}$ & \faVolumeDown \\
    Singer Attribution & Generate a jazz piece accompanied by lyrics "<lyrics>", featuring a tenor singer performing in Bel Canto style. & $\mathbb{T}$ & 
    \faVolumeDown \\
    Lyrics Editing & Replace the lyrics in <audio\_0> with <lyrics>, keeping the original melody unchanged. & $\mathbb{T}$, \faVolumeDown & \faVolumeDown \\
    Transition Visualization & Generate three images showing the transition process from <image\_0> to <image\_1>. & $\mathbb{T}$, \faImages & \faImages \\
    Future Prediction & Generate three images showing the future events after <image\_0>. & $\mathbb{T}$, \faImage & \faImages \\
    Speech Encoding & Generate a speech about sustainable development, and provide the speech transcript encoded in Base64. & $\mathbb{T}$ & $\mathbb{T}$, \faVolumeDown \\
    Image-to-Sound & Create a music predominately featuring the instrument shown in <image\_0>. & $\mathbb{T}$, \faImage & \faVolumeDown \\
    Sound-to-Image & Draw an image showing the animal that is mostly likely to make the sound in <audio\_0>. & $\mathbb{T}$, \faVolumeDown & \faImage \\
    \bottomrule[1.5pt]
    \end{tabular}
\vspace{5pt}
\caption{Tasks that are not included in \textsc{MMMG}. $\mathbb{T}$ denotes text modality, \faImage \ for image modality, \faImages \ for multiple images, \faVolumeDown \ for audio and \faVolumeDowns \ for multiple audios. We hope to incorporate these tasks when reliable evaluation methods are available.}
\label{tab:other_task}
\end{table}
\subsection{Dataset Statistics}

We present some important statistics of \textsc{MMMG} in Table \ref{tab:stat}.

\subsection{Computation Statistics}
\label{sec:compute}
\begin{wraptable}{r}{0.5\linewidth}{
\centering
\resizebox{1\linewidth}{!}{
\scriptsize
\begin{tabular}{ll}\toprule[1.5pt]
\textbf{Statistics} &\textbf{Number} \\
\midrule[0.75pt]
Total number of modality combinations &4\\
Total number of tasks &55 \\
\quad - I : A : I-T : A-T & 15 : 12 : 22 : 6\\
Total number of questions &1288 \\
\quad - I : A : I-T : A-T & 410 : 238 : 480 : 160\\
Total number of images &542 \\
Total number of audios &90 \\
Average length of instructions & 242.5\\
\bottomrule[1.5pt]
\end{tabular}}
\caption{Statistics of \textsc{MMMG}. I, A, I-T, A-T stands for image, audio, interleaved image-text and interleaved audio-text generations respectively.}
\label{tab:stat}
}
\end{wraptable}
The evaluation pipeline for \textsc{MMMG} requires at least a single NVIDIA A10 GPU for open-source models, and APIs from OpenAI and Gemini for proprietary models. In our experiments, we used a single NVIDIA A40 GPU. On average, the evaluation runtime for each task is approximately 4 minutes, incurring a API cost of about \$1.1 for a sample size of 4. For the generation phase, runtime significantly varies depending on the model itself. The most time-consuming model tested is \textsc{Yue}, which runs on a single NVIDIA H100 GPU. On average, \textsc{Yue} takes around 3 hours to complete generation per task.

\subsection{Design Principle}
For single-modality generation tasks (image, audio, speech), we follow established task collections from prior work that have identified critical capabilities including object composition, spatial reasoning, attribute control, text rendering and instruction following, etc. \citep{ghosh2023geneval, li2024genai}. However, for interleaved multimodal generation, no systematic guidelines exist for determining which tasks are most important for comprehensive evaluation.
To address this gap, we systematically identify three fundamental capabilities for interleaved multimodal generation based on analysis of prior work and real-world applications \citep{chen2024interleaved, xia2024mmie, deng2025emerging}: (1) \textbf{Input preservation and understanding}: The ability to accurately process and retain information from multimodal inputs, including understanding relationships between provided images, text, and their combinations. (2) \textbf{Modality sequence consistency}: The ability to generate coherent sequences of outputs (e.g., multiple images or speech segments) that maintain temporal, spatial or semantic consistency. (3) \textbf{Cross-modal coherence}: The ability to generate outputs where different modalities (text and images, text and audio) are semantically aligned and mutually supportive and also keep modality number and order correct.

These capabilities are assessed through carefully selected proxy tasks. For instance, image editing tasks serve as diagnostic measures for input preservation and understanding. A model that cannot accurately modify specific objects while preserving unchanged regions likely lacks the fine-grained multimodal comprehension necessary for more complex interleaved generation tasks. This design is supported by recent findings that during interleaved multimodal pre-training, models develop basic multimodal understanding and generation abilities before complex editing and reasoning capabilities emerge \citep{deng2025emerging}, establishing a developmental hierarchy where foundational skills are prerequisites for higher-order reasoning.

Our empirical results validate this hierarchical assumption: current models already struggle significantly with basic object-focused editing tasks (Section \ref{sec:exp_benchmark}), while complex reasoning tasks in math and code show near-zero performance across most models. This pattern suggests that current limitations in complex compositional reasoning may stem from deficiencies in foundational capabilities. As more advanced models emerge with improved capabilities, we plan to expand MMMG's task scope to include increasingly complex compositional reasoning tasks, enabling deeper investigation of the relationships between foundational and advanced capabilities.

\section{Detailed Experiment Setup}
\label{sec:detail_exp}
\subsection{Model Details}
\label{sec:detail_model}

\paragraph{Generation.} We evaluate 29 multimodal generation models specified in Appendix \ref{sec:detail_model}. To encourage diversity, we only incorporate the latest model of a series. Even though our benchmark supports comprehensive and cross-modality evaluation, current multimodal generation models have very restricted output modalities. Thus, we categorize these models by their supported output modalities into image, interleaved image-text, sound-music, and interleaved speech-text generation.
\begin{itemize}[leftmargin=*]
    \item \textbf{Image Generation.} We include \textsc{GPT Image} \citep{openai2024gptimage}, through the ``gpt-image-1'' checkpoint on OpenAI API; \textsc{Imagen 3} \citep{baldridge2024imagen}, through the ``imagen-3.0-generate-002'' checkpoint on Gemini API; \textsc{Recraft V3} \citep{Recraft2024}, through the ''recraftv3'' checkpoint on Recraft API; \textsc{Luma Photon} \citep{Luma2024}, through the ``luma/photon'' checkpoint on Replicate API; \textsc{Flux 1.1 Pro} \citep{flux2024}, through the ``black-forest-labs/flux-1.1-pro'' checkpoint on Replicate API; \textsc{Ideogram 2} \citep{Ideogram2024}, through the ``ideogram-ai/ideogram-v2'' checkpoint on Replicate API; \textsc{Dalle 3} \citep{betker2023improving}, through the ``dall-e-3'' checkpoint on OpenAI API; \textsc{Stable Diffusion 3.5} \cite{rombach2022high}, through the ``stabilityai/stable-diffusion-3.5-large'' checkpoint on Huggingface; \textsc{Janus Pro} \citep{chen2025janus}, through \href{https://github.com/deepseek-ai/Janus}{the official implementation}; \textsc{BLIP-3o} \citep{chen2025blip3}, through \href{https://github.com/JiuhaiChen/BLIP3o}{the official implementation}.
    \item \textbf{Interleaved Image-Text Generation.} We include \textsc{SEED-LLaMa} \citep{ge2023making}, through \href{https://github.com/AILab-CVC/SEED}{the official implementation}; \textsc{Anole} \citep{chern2024anole}, through \href{https://github.com/GAIR-NLP/anole}{the official implementation} on Github; \textsc{Gemini 2 Image} \citep{team2023gemini}, through the ``gemini-2.0-flash-preview-image-generation'' checkpoint on Gemini API; \textsc{Gemini 2.5 Image} \citep{team2023gemini}, through the ``gemini-2.5-flash-image-preview'' checkpoint on Gemini API; and \textsc{GPT Image} \citep{team2023gemini}, through the ``gpt-image-1'' checkpoint on OpenAI API for image generation and the ``gpt-4.1'' with ``image\_generation'' tool on OpenAI API for interleaved image-text generation. We also implement two agents models composing of a MLM and an image generation model: \textsc{GPT-4o} + \textsc{GPT Image} and \textsc{Gemini 2.5} + \textsc{GPT Image}. \textsc{Gemini 2.5} is through the ``gemini-2.5-pro-preview-03-25'' checkpoint on Gemini API and \textsc{GPT-4o} is through the ``gpt-4o-2024-08-06'' checkpoint on Openai API.
    \item \textbf{Sound and Music Generation.} We include \textsc{Stable Audio} \citep{evans2025stable}, through the ``stabilityai/stable-audio-open-1.0'' checkpoint on Huggingface, and  \textsc{AudioLDM 2} \citep{liu2024audioldm}, through the ``cvssp/audioldm2-large'' checkpoint on Huggingface, capable of generating both sound and music. We also include sound generation models: \textsc{AudioGen} \citep{kreuk2022audiogen}, through \href{https://github.com/facebookresearch/audiocraft/blob/main/docs/AUDIOGEN.md}{the official implementation}; \textsc{Make-An-Audio 2} \citep{huang2023make}, through \href{https://github.com/bytedance/Make-An-Audio-2}{the official implementation}; and \textsc{Tango 2} \citep{majumder2024tango}, through the ``declare-lab/tango2-full'' checkpoint on Huggingface. We also include music generation models: \textsc{MusicGen} \citep{copet2023simple}, through the ``facebook/musicgen-large'' checkpoint on Huggingface; \textsc{Tango Music} \citep{kong2024improving}, through the ``declare-lab/tango-music-af-ft-mc'' checkpoint on Huggingface; and \textsc{Yue} \citep{yuan2025yue}, through \href{https://github.com/multimodal-art-projection/YuE}{the official implementation}.
    % yue guidance and lyrics seperation
    \item \textbf{Interleaved Speech-Text Generation.} We include \textsc{Spirit LM} \citep{nguyen2025spirit}, through \href{https://github.com/facebookresearch/spiritlm}{the official implementation}. We also implement two agents models composing of a MLM and a voice synthesizing model: \textsc{Gemini 2.5} + \textsc{VoxInstruct} \citep{zhou2024voxinstruct} and \textsc{Gemini 2.5} + \textsc{VoiceLDM} \citep{lee2024voiceldm}. \textsc{VoxInstruction} is through \href{https://github.com/thuhcsi/VoxInstruct}{the official implementation} and \textsc{VoiceLDM} is through \href{https://github.com/glory20h/VoiceLDM}{the official implementation}.
\end{itemize}
\textbf{Evaluation.} We compare several evaluation methods. For image generation, we include \textsc{GPT-4o}, \textsc{Gemini 2.5}, and \textsc{Qwen2.5-VL-72B-Instruct} \citep{bai2025qwen2} to perform VQA for evaluation. $\mathrm{CLIPScore}$ \citep{hessel2021clipscore} is found as less aligned with human judgment in previous studies \citep{hu2023tifa}, thus not included. For sound and music evaluation, we include $\mathrm{CLAPScore}_{\mathrm{audio}}$, $\mathrm{CLAPScore}_{\mathrm{text}}$, and employing \textsc{Gemini 2.5} for acoustic question answering (AQA). $\mathrm{CLAPScore}_{\mathrm{audio}}$ computes the CLAP cosine similarity with reference audio, while $\mathrm{CLAPScore}_{\mathrm{text}}$ computes the similarity with reference audio captions. We pick the the optimal thresholds separately per dataset.

The checkpoints of the evaluation models we use are: \textsc{GPT-4o}, through the ``chatgpt-4o-latest'' checkpoint on OpenAI API; \textsc{Gemini 2.5}, through the ``gemini-2.5-pro'' checkpoint on Gemini API; and \textsc{QWEN2.5-VL}, through the ``Qwen/Qwen2.5-VL-72B-Instruct'' checkpoint on Huggingface. For audio models, we employ \textsc{CLAP}, through the ``laion/clap-htsat-unfused'' checkpoint on Huggingface; \textsc{Whisper}, through the ``openai/whisper-large-v3'' checkpoint on Huggingface and a finetuned Chinese speech-to-text checkpoint ``BELLE-2/Belle-whisper-large-v3-zh'' on Huggingface; \textsc{WavLM}, through the ``microsoft/wavlm-base-sv'' checkpoint on Huggingface; and ``Wav2Vec'', through the ``alefiury/wav2vec2-large-xlsr-53-gender-recognition-librispeech'' checkpoint on Huggingface.
\subsection{Generation Details}
Following the experimental setup in \citet{ghosh2023geneval}, we sample 4 generations for every instruction in our benchmark. We employ a temperature of 0 and a retry count of 4 for MLMs and sampling steps of 200 for diffusion models. We keep other parameters, such as guidance scale, as default values. For non-agent models, we directly provide instructions to the model. For agent-based models, we prepend a system prompt to the instructions. This system prompt explicitly instructs the model to generate outputs following a structured, function-call-based approach. When the model needs visual or auditory outputs, it generates placeholders formatted as function calls within the text. Each placeholder clearly specifies the generation instructions and any necessary references to prior outputs or provided multimedia in user's instructions. For each placeholder, we extract the function call, which are then fed into specialized image or audio generation models. To correctly handle references to previously generated media, we employ topological sorting. This ensures media outputs are generated in a sequence by dependencies, and circular dependencies are identified and reported as errors. Detailed system prompt for interleaved image-text agent is in Table \ref{tab:i_agent_prompt} and interleaved audio-text agent is in Table \ref{tab:a_agent_prompt}.
\begin{table}[!ht]
\centering
% \scriptsize
\begin{tabular}{p{0.95\linewidth}}
\toprule[1.5pt]
You are a multimodal assistant capable of generating both text and images. When visual content would enhance your response or is specifically requested, you can generate or edit images through advanced diffusion models.

To generate or edit an image:

\begin{enumerate}
\item Identify when visual content would be beneficial or requested.
\item Insert an image generation/editing placeholder using the following format:
\begin{quote}
\texttt{\textless image\_start\textgreater\textless image\_prompt="Detailed image generation or editing prompt here."\textgreater\textless image\_ref=[reference identifiers]\textgreater\textless image\_end\textgreater}
\end{quote}
\item The post-processing system replaces this placeholder with an image created or edited based on your instructions.
\item Naturally incorporate references to the generated or edited image in your ongoing conversation.
\end{enumerate}

When crafting image prompts, follow these guidelines:

\textbf{For image prompts:}

\begin{itemize}
\item Provide detailed, specific descriptions (15-30 words) for optimal results.
\item Include artistic styles (photorealistic, cartoon, watercolor, etc.) or style transfers.
\item Specify key objects and their attributes (colors, textures, etc.), or modifications.
\item Detail composition elements (spatial relationships, perspective, lighting, etc.), or compositional changes.
\item Ensure instructions are clear and concise.
\end{itemize}

\textbf{For image references:}

Three reference types are available:

\begin{enumerate}
\item Image generation (no reference):
\begin{quote}
\texttt{<image\_ref=[]>}
\end{quote}

\item Editing user-provided images:
\begin{quote}
    Format: \texttt{<image\_ref=[i]>} where \texttt{i} is the index of the provided image (indices starting at 0).

    Example: \texttt{<image\_ref=[0]>} references the first provided image.

    Multiple images example: \texttt{<image\_ref=[0,2]>} references the first and third provided images.
\end{quote}

\item Editing previously generated images:
\begin{quote}
    Format: \texttt{<image\_ref=[\#N]>}, where \texttt{N} is the sequential number of previously generated images (starting from 0).

    Example: \texttt{<image\_ref=[\#3]>} references the fourth generated image.

    Multiple images example: \texttt{<image\_ref=[\#0,\#2]>} references the first and third generated images.
\end{quote}

\end{enumerate}

\textbf{Important:} Use only one reference type within each placeholder. Different reference types may be used across multiple placeholders.

Provide concise and direct responses following user instructions precisely. Always maintain the exact placeholder format for proper parsing, ensuring that both images and text appear in the required order. Do not omit any necessary text following image placeholders.
\\
\bottomrule[1.5pt]
\end{tabular}
\vspace{5pt}
\caption{System prompt for interleaved image-text agent.} \label{tab:i_agent_prompt}
\end{table}
\begin{table}[!ht]
\centering
% \scriptsize
\begin{tabular}{p{0.95\linewidth}}
\toprule[1.5pt]

You are a multimodal assistant capable of generating both text and audio. When audio content would enhance your response or is specifically requested, you can generate audio through text-to-audio models.

To generate audio:

\begin{enumerate}
\item Identify when audio content would be beneficial or requested.
\item Insert an audio generation placeholder using the format:
\begin{quote}
\texttt{\textless audio\_start\textgreater\textless audio\_type="sound" OR "speech" OR "music"\textgreater\textless audio\_text="Text to be spoken here."\textgreater\textless audio\_style="Descriptive text here." OR audio reference ID\textgreater\textless audio\_end\textgreater}
\end{quote}
\item The post-processing system replaces this placeholder with generated audio based on your specifications.
\item Naturally incorporate references to the generated audio in your ongoing conversation.
\end{enumerate}

When crafting audio prompts, follow these guidelines:

\textbf{Audio Type:}
\begin{itemize}
\item Must be exactly one of: \texttt{"sound"}, \texttt{"speech"}, or \texttt{"music"}.
\item \texttt{"speech"}: For human speech.
\item \texttt{"sound"}: For environmental sounds or effects.
\item \texttt{"music"}: For musical compositions or instrumental pieces.
\end{itemize}

\textbf{Audio Text:}
\begin{itemize}
\item For \texttt{"speech"}: Provide the exact transcript.
\item For \texttt{"sound"} or \texttt{"music"}: Leave as empty string ("").
\item Keep speech concise (typically under 50 words).
\end{itemize}

\textbf{Audio Style:}
\begin{enumerate}
\item \textbf{Descriptive Text:}
\begin{itemize}
\item For \texttt{"speech"}: Specify voice characteristics (gender, emotion, pace, pitch, accent).
\item For \texttt{"sound"}: Specify sound source, environment, qualities.
\item For \texttt{"music"}: Specify genre, mood, tempo, instruments.
\end{itemize}
\item \textbf{Reference Audio:}
\begin{itemize}
\item For consistency, particularly with speech:
\begin{itemize}
\item Previously generated audio: \texttt{\textless audio\_style=\#N\textgreater} (\texttt{N} is sequential number starting at 0).
\item User-provided audio: \texttt{\textless audio\_style=N\textgreater} (\texttt{N} is sequential number of provided audio starting at 0).
\end{itemize}
\item Important: Only reference audio that itself does not reference previous audio to avoid circular references.
\end{itemize}
\end{enumerate}

Provide concise, direct responses precisely following user instructions. In multi-speaker scenarios, maintain consistent and distinctive voice characteristics for each speaker. Always maintain the exact placeholder format for correct parsing
\\
\bottomrule[1.5pt]
\end{tabular}
\vspace{5pt}
\caption{System prompt for interleaved audio-text agent.} \label{tab:a_agent_prompt}
\end{table}
\subsection{Evaluation Details}
\label{sec:detail_eval}
\paragraph{Prompts for VLMs}
\begin{itemize}[leftmargin=*]
    \item \textbf{Object Count.} ``\emph{How many [object] are there in the given image? Choose from the options: A. Less than 3 or the image is blank B. 3 C. 4 D. 5 E. 6 F. More than 6. Respond only with the option letter (A, B, C, D, E or F). Do not provide any explanation, reasoning, or additional information.}'' Multiple choice questions can boost VLM's performance on object count tasks. We employ this prompt for object count and self count tasks.
    \item \textbf{Absolute Spacial Relationship.} ``\emph{The [object] is located in which section of the image? Choose from the options: A. bottom left B. bottom right C. up left D. up right E. none of the above (positioned in a more central way) Explain step by step and end your answer with Answer: [only an optional letter].}'' Multiple choice questions can boost VLM’s performance on spacial reasoning tasks. We employ this prompt for absolute spatial relationship and self absolute spatial relationship recognizing tasks. 
    \item \textbf{Left-Right Spacial Relationship.} ``\emph{Looking at the 2D composition of the image, what is the horizontal alignment relationship between the [object1] and the [object2]? Choose from the options: A. the [object1] is obviously to the left of the [object2]. B. the [object1] is obviously to the right of the [object2]. C. the [object1] is neither obviously to the right nor left of the [object2]. Explain step by step and end your answer with Answer: [only an optional letter].}'' VLMs tend to be confused by perspective relationship, thus we ask VLMs to focus on 2D composition. We employ this prompt for relative spatial relationship and self relative spatial relationship recognizing tasks.
    \item \textbf{Up-Down Spacial Relationship.} ``\emph{Looking at the 2D composition of the image, what is the vertical alignment relationship between the [object1] and the [object2]? Choose from the options: A. the [object1] is obviously positioned higher than the [object2]. B. the [object1] is obviously positioned lower than the [object2]. C. the [object1] is neither obviously positioned higher nor lower than the [object2]. Explain step by step and end your answer with Answer: [only an optional letter].}'' We employ this prompt for relative spatial relationship and self relative spatial relationship recognizing tasks.
    \item \textbf{OCR English.} ``\emph{\#\#\# Instruction: Recognize all the major texts (ignore small texts on the edge) ONLY on [object]. Only recognize texts in Latin alphabet characters (a-z, A-Z). Do not correct the text if it is misspelled, nonsense or wrong, output the most direct recognition result. Do not call any function. \#\#\# Output format: Output an executable Python list of all recognized texts from top to down, from left to right, e.g. [``Hello World'', ``Good morning'']. Output an empty list if the there is no text on [object] or the image is blank.}'' We employ this prompt for single and double text rendering and self OCR tasks.
    \item \textbf{OCR Chinese.} ``\emph{\#\#\# Instruction: You are a conservative text recognition model. Your task is to recognize all the major Chinese characters in the given image. If the Chinese characters in the image are wrongly written or distorted, you should return an empty string. Do not call any function. \#\#\# Output format: Only a string of all recognized characters from top to down, from left to right. Do not add quotations.}'' We employ this prompt for multi-lingual text rendering task. Since VLMs tend to recognize Chinese characters incorrectly or identify fake characters, we employ two separate VLMs and use the intersection of their recognition results to improve accuracy.
    \item \textbf{Text Pattern Verifying (Math)} ``\emph{Below are two descriptions of the same geometric pattern, one is ground-truth and the other is model-generated. Your task is to judge if the generated description is accurate. Analyze step by step and end your answer with ``Yes'' or ``No''. Here are some criteria: 1. The model-generated pattern must state the pattern clearly without ambiguity. For example, a 3*3 grid of circles with some circles filled is ambiguous. 2. Make sure the overall structure, the position and situation of each element are accurate. Specifically, the situation of each element can include: filled (black, grey, filled with black or any equivalent words), unfilled (white, hollow, empty or any equivalent words), missing (the position is empty or missing). If the situation is not specified in the ground-truth, the element can take any situation of the right shape. 3. If the ground-truth describes a coordinate system, the x-axis will increase from left to right while y-axis will increase from top to down. For example, for a 3*3 grid, the (3,2) coordinate is the middle-right element.}'' We employ this prompt for math task.
    \item \textbf{Image Verifying (Math)} ``\emph{Your task is to judge if the given image accurately follows the ground-truth pattern. Analyze step by step and end your answer with ``Yes'' or ``No''. Here are some criteria: 1. Make sure the overall structure, the position and situation of each element are accurate. Specifically, the situation of each element can include: filled (black, grey, filled with black or any equivalents), unfilled (white, hollow, empty or any equivalents), missing (the position is empty, missing or any equivalents). If the situation is not specified in the ground-truth, the element can take any situation of the right shape. 2. If the ground-truth describes a coordinate system, the x-axis will increase from left to right while y-axis will increase from top to down. For example, for a 3*3 grid, the (3,2) coordinate is the middle-right element. 3. If the given image contains multiple patterns (e.g. multiple grids) or question mark, the given image doesn't follow the ground-truth pattern.}'' We employ this prompt for math task.
    \item \textbf{Object Existing.} ``\emph{Is/Are there [detailed object description] in the given image? Explain step by step and end your answer with ``Yes'' or ``No''. Answer ``No'' if the image is blank.}'' We design detailed object description for each instruction manually, include object number, object attributes and undesired negative attributes, etc.. We employ this prompts for all image tasks unmentioned above. For spatial relation tasks, we first exam if the object number is accurate by object existing prompt and then check spatial relationship by corresponding prompts.
\end{itemize}
\paragraph{Program Verifying}
\begin{itemize}[leftmargin=*]
    \item \textbf{Solid Color Fill.} The evaluation procedure starts by cropping the targeted region from the image and calculating its average RGB value. The average RGB value is compared with a standard reference color; if the relative deviation exceeds 15\%, indicating significant color discrepancy, the evaluation returns zero. Next, structural consistency is assessed by computing the SSIM between the targeted region and an artificially generated solid region filled with the calculated average RGB color, confirming color uniformity. Finally, the procedure examines over-fill by evaluating the margin area surrounding the targeted region and computing the proportion of pixels matching the region's average RGB color. The ratio as penalty is subtracted from the SSIM score.
    \item \textbf{Image Editing.} The evaluation for image editing begins by manually labeling a potential editing area within each image. Then crop the edited area from the generated image and compare against the corresponding area in a reference image or assessed via a VLM. Additionally, regions outside this area are compared with corresponding original outside area using SSIM to detect unintended changes. The final score is the product of these two comparisons, reflecting editing accuracy and preservation of original content.
    \item \textbf{Sound Generation.} For begin-end tasks, clip the first or last 4 seconds of audio directly. For positional inclusion tasks, crop the corresponding fraction of the audio. For silence detection tasks, utilize the \texttt{librosa.effects.split} function to segment audio based on silence intervals and then verify if each section contains target sound through $\mathrm{CLAPScore}_{\mathrm{udio}}$.
    \item \textbf{Music Generation.} For tempo evaluation, use \textsc{BeatThis} to extract beat tracks and calculate Beats Per Minute (BPM). For intensity evaluation, analyze the initial and final 4 seconds of the music, plotting the energy spectrum through \texttt{librosa.feature.rms} and computing its slope and goodness of fit. Only audio segments demonstrating clear upward or downward trends in energy pass the intensity evaluation.
    \item \textbf{Speech Generation.} For pitch evaluation, calculate the average energy of each pitch through \texttt{parselmouth.Sound.to\_pitch} and select the pitch with the highest average energy through \texttt{parselmouth.Sound.to\_intensity} as the speech pitch. For speed evaluation, transcribe English audio using \textsc{Whisper} and compute words per minute (WPM); for Chinese audio, compute characters per minute (CPM). For textual constraints, normalize transcripts using  \textsc{Whisper}'s tokenizer (removing punctuation, case sensitivity, etc.) and evaluate with the tools of IFEval. For speech translation task, we calculate BLEU \citep{papineni2002bleu} and for other speech tasks, we calculate Word Accuracy.
\end{itemize}
\subsection{Annotation Interface}
\begin{figure}[!t]
    \centering
    \includegraphics[width=1\linewidth]{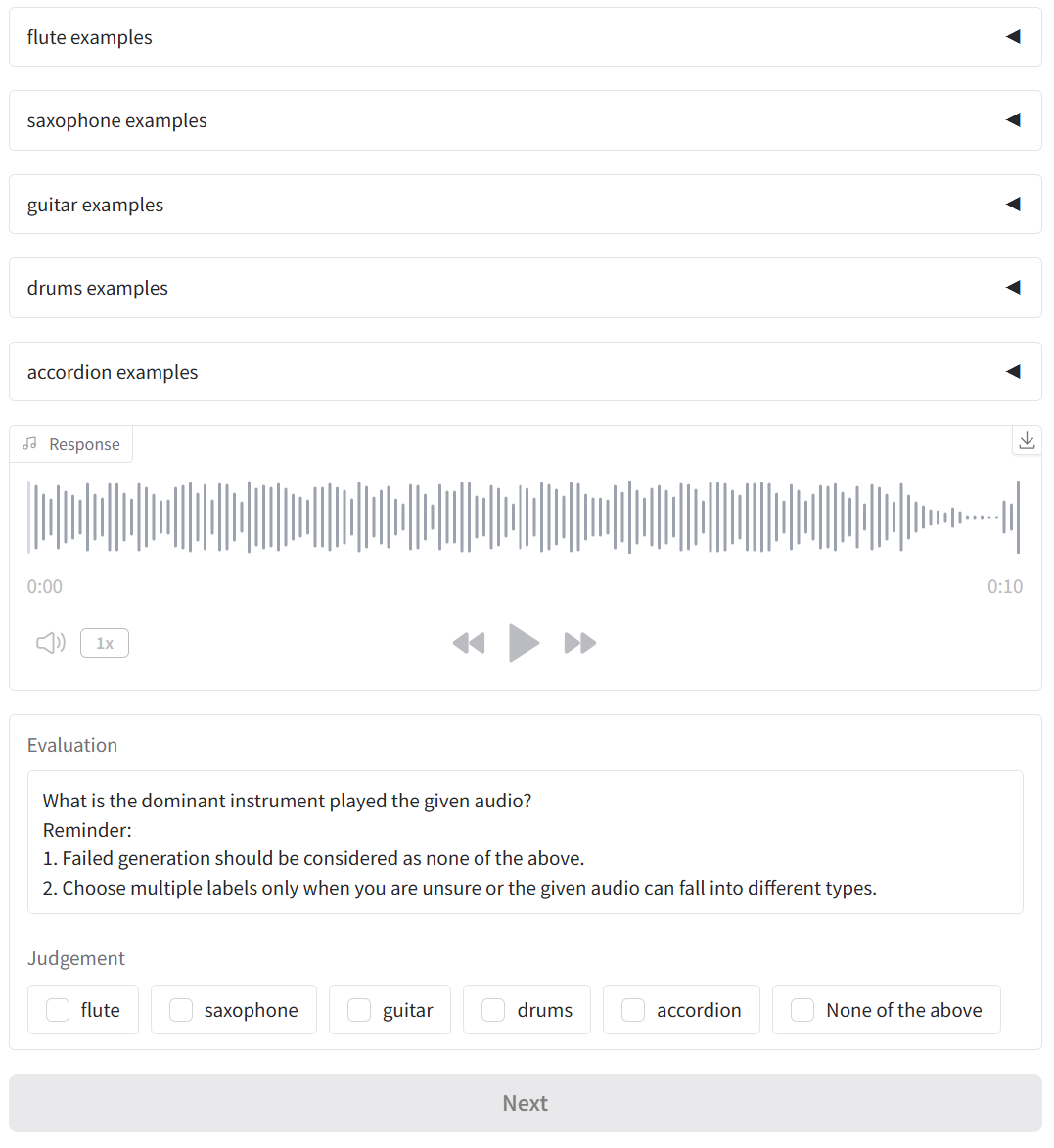}
    \caption{Human annotation interface for instrument inclusion task. Typically, an inference will include reference audios/images, model's generation, evaluation instruction, evaluation criteria and judgment radio boxes and next/previous button.}
    \label{fig:interface}
    \vspace{-15pt}
\end{figure}
We design task-specific annotation interfaces by Gradio \citep{abid2019gradio}, each including reference images or audio, model's generated outputs, judgment instructions, and judgment criteria. We preprocess some generated outputs to assist annotators in their judgments. For example, we provide cropped images within editing area for image editing tasks and clipped audio segments at the beginning or end for audio begin-end tasks. Judgments are typically collected through multiple-choice radio buttons to ensure high inter-annotator agreement. However, for OCR tasks specifically, annotators type the recognized text directly. An example of annotation interface is in Figure \ref{fig:interface}.

\subsection{Annotation Questions}
\label{sec:human_question}
For image and interleaved image-text evaluation tasks, we employ the same questions as the prompts used for VLMs. We paraphrase the questions to make them more annotator friendly and add judging criteria to reduce the ambiguity of the questions. For audio and interleaved audio-text evaluation tasks, we design new annotations questions as follow:
\paragraph{Music Instrument.} ``\emph{What is the dominant instrument played the given audio? Reminder: 1. Failed generation should be considered as none of the above. 2. Choose multiple labels only when you are unsure or the given audio clearly have different types of instruments.}'' We employ this question for instrument inclusion and exclusion tasks.
\paragraph{Music Genre.} ``\emph{What is the dominant genre played the given audio? Reminder: 1. Failed generation should be considered as none of the above. 2. Choose multiple labels only when you are unsure or the given audio can fall into different types. 3. Only choose a genre when it is very obvious/typical.}'' We employ this question for instrument inclusion and exclusion tasks.
\paragraph{Sound Inclusion.}  ``\emph{Is the given audio about [sound]? Reminder: 1. Chose yes when [sound] is the main sound existing in the audio. 2. [sound] should be common real-world sound without distortion.}'' We employ this question for all sound generation tasks.
\paragraph{Speaker Similarity.}  ``\emph{Are the speeches coming from the same speaker? Reminder: 1. Little speaker voice difference can be tolerated, but overall, there should be no major difference.}'' We employ this question for voice replication and conversation tasks.
\paragraph{Speaker Gender}  ``\emph{What is the gender of the speaker in the given speech? Reminder: 1. Choose none of above when the voice sounds like electronic synthesizer sound or it is hard to categorize into binary genders. 2. Do not consider speech quality (clarity and fluency, etc.) when judging gender.}'' We employ this question for voice attribution and multi-lingual speech tasks.
\section{Experiment Results (Cont.)}

\subsection{Correlation with Human Annotation}
\label{sec:corr}
\begin{table}[!ht]
\centering
\scriptsize
\tabcolsep=0.02cm
\begin{tabular}{l>{\centering\arraybackslash}m{0.1\textwidth}>{\centering\arraybackslash}m{0.1\textwidth}>{\centering\arraybackslash}m{0.1\textwidth}>{\centering\arraybackslash}m{0.1\textwidth}>{\centering\arraybackslash}m{0.1\textwidth}>{\centering\arraybackslash}m{0.1\textwidth}>{\centering\arraybackslash}m{0.1\textwidth}>{\centering\arraybackslash}m{0.1\textwidth}}\toprule[1.5pt]
\multirow{2}{*}{\textbf{Task}} &\multicolumn{2}{c}{\textsc{GPT-4o}} &\multicolumn{2}{c}{\textsc{Gemini 2.5}} &\multicolumn{2}{c}{\textsc{Qwen2.5-VL}} & \multicolumn{2}{c}{IAA} \\
& agree & corr & agree & corr & agree & corr & agree & corr\\
\midrule[0.75pt]
Object Inclusion &\textbf{0.925} &\textbf{0.776} &0.900 &0.715 &0.888 &0.657 &1.000 &1.000 \\
Object Exclusion &\textbf{0.963} &\textbf{0.924} &0.913 &0.823 &0.925 &0.855 &1.000 &1.000 \\
Object Count &0.875 &0.709 &0.963 &\textbf{0.912} &\textbf{0.900} &0.763 &0.975 &0.943 \\
Object Knowledge &\textbf{0.963} &\textbf{0.925} &\textbf{0.963} &\textbf{0.925} &0.938 &0.875 &1.000 &1.000 \\
Object Commonsense &0.913 &0.787 &0.888 &0.696 &\textbf{0.938} &\textbf{0.822} &1.000 &1.000 \\
Object Attribution &\textbf{0.938} &\textbf{0.848} &\textbf{0.938} &0.835 &0.900 &0.758 &1.000 &1.000 \\
Compassion Relation &\textbf{0.925} &\textbf{0.850} &0.875 &0.741 &0.850 &0.699 &0.950 &0.896 \\
Universal Relation &\textbf{0.975} &\textbf{0.951} &0.900 &0.818 &0.925 &0.860 &0.975 &0.951 \\
Relative Spatial &\textbf{0.925} &\textbf{0.819} &0.825 &0.640 &0.800 &0.572 &0.950 &0.875 \\
Absolute Spatial &0.825 &0.641 &\textbf{0.925} &\textbf{0.839} &0.839 &0.775 &0.983 &0.960 \\
Text Rendering (TR) &0.991 &0.994 &\textbf{0.992} &\textbf{1.000} &0.967 &0.942 &1.000 &1.000 \\
Double TR &\textbf{0.841} &\textbf{0.906} &0.646 &0.662 &0.574 &0.471 &0.938 &0.938 \\
Multi-lingual TR &\textbf{0.889} &\textbf{0.989} &\textbf{0.889} &0.968 &0.800 &0.875 &1.000 &1.000 \\
Semantic &\textbf{0.958} &\textbf{0.910} &0.946 &0.890 &0.940 &0.869 &0.982 &0.961 \\
Composition &\textbf{0.971} &\textbf{0.930} &0.942 &0.847 &0.920 &0.793 &0.978 &0.944 \\
Decomposition &\textbf{0.971} &\textbf{0.941} &\textbf{0.971} &\textbf{0.941} &0.949 &0.900 &0.978 &0.956 \\
Text Adding &\textbf{0.969} &\textbf{0.996} &0.750 &0.710 &0.927 &0.980 &0.950 &0.950 \\
Text Altering &0.950 &1.000 &0.950 &0.976 &\textbf{0.975} &\textbf{1.000} &0.950 &0.950 \\
Object Adding &\textbf{0.975} &\textbf{0.912} &0.925 &0.728 &0.825 &0.428 &1.000 &1.000 \\
Object Removing &0.975 &0.933 &0.975 &0.933 &\textbf{1.000} &\textbf{1.000} &1.000 &1.000 \\
Object Replacing &0.925 &0.819 &\textbf{0.975} &\textbf{0.941} &0.900 &0.762 &0.925 &0.819 \\
Object Altering &\textbf{0.975} &\textbf{0.951} &0.875 &0.747 &0.850 &0.704 &0.925 &0.819 \\
Self Count &\textbf{0.975} &\textbf{0.950} &0.950 &0.899 &0.525 &0.006 &1.000 &1.000 \\
Self Color &\textbf{0.950} &0.881 &\textbf{0.950} &\textbf{0.883} &0.642 &0.363 &0.983 &0.960 \\
Self Size &\textbf{0.892} &\textbf{0.788} &0.867 &0.735 &0.917 &0.838 &0.967 &0.933 \\
Self OCR &\textbf{0.906} &\textbf{0.909} &0.806 &0.790 &0.748 &0.690 &1.000 &1.000 \\
Self Relative Spatial &0.838 &0.669 &\textbf{0.950} &\textbf{0.896} &0.813 &0.605 &0.963 &0.923 \\
Self Absolute Spatial &0.913 &0.821 &\textbf{0.950} &\textbf{0.897} &0.963 &0.923 &0.975 &0.948 \\
Math &0.950 &0.436 &\textbf{1.000} &\textbf{1.000} &0.950 &-0.026 &0.988 &0.703 \\
Code &\textbf{1.000} &\textbf{1.000} &\textbf{1.000} &\textbf{1.000} &\textbf{1.000} &\textbf{1.000} &1.000 &1.000 \\
\midrule[0.75pt]
Average &\textbf{0.935} &\textbf{0.865} &0.913 &0.846 &0.867 &0.718 &0.980 &0.950 \\
\bottomrule[1pt] \\

\toprule[1pt] 
\multirow{2}{*}{\textbf{Task}} & \multicolumn{2}{c}{$\mathrm{CLAPScore_{\mathrm{audio}}}$} &\multicolumn{2}{c}{$\mathrm{CLAPScore_{\mathrm{text}}}$} &\multicolumn{2}{c}{\textsc{Gemini 2.5}} & \multicolumn{2}{c}{IAA}\\
& agree & corr & agree & corr & agree & corr & agree & corr\\
\midrule[0.75pt]
Sound Begin-End &\textbf{0.925} &\textbf{0.951} &0.825 &0.687 &0.625 &0.204 &0.967 &0.933 \\
Sound Inclusion &\textbf{0.850} &\textbf{0.711} &0.800 &0.564 &0.650 &0.207 &0.925 &0.856 \\
Sound Knowledge &\textbf{0.944} &\textbf{0.817} &0.861 &0.534 &0.639 &0.439 &0.917 &0.720 \\
Sound Silence &\textbf{0.975} &\textbf{0.946} &\textbf{0.975} &\textbf{0.946} &0.950 &0.690 &1.000 &1.000 \\
Instrument Inclusion &\textbf{0.967} &\textbf{0.894} &\textbf{0.967} &\textbf{0.894} &0.967 &0.894 &1.000 &1.000 \\
Instrument Exclusion &\textbf{0.893} &\textbf{0.663} &0.325 &0.189 &0.825 &0.378 &0.929 &0.782 \\
Music Genre &\textbf{0.900} &\textbf{0.764} &0.775 &0.435 &0.750 &0.355 &0.929 &0.782 \\
\midrule[0.75pt]
Average &\textbf{0.923} &\textbf{0.821} &0.790 &0.607 &0.772 &0.452 &0.956 &0.882 \\
\bottomrule[1pt] \\ 

\toprule[1pt]
\multirow{2}{*}{\textbf{Task}} & \multicolumn{2}{c}{WavLM} &\multicolumn{2}{c}{Wav2Vec} & \multicolumn{2}{c}{IAA}\\
& agree & corr & agree & corr & agree & corr\\
\midrule[0.75pt]
Voice Attribution &- &- &\textbf{0.949} &\textbf{0.826} &0.950 &0.844 \\
Voice Replication &\textbf{0.875} &\textbf{0.731} &- &- &0.925 &0.843 \\
Speech Multi-lingual &- &- &\textbf{0.966} &\textbf{0.876} &0.925 &0.856 \\
Conversation &\textbf{0.850} &\textbf{0.630} &- &- &0.925 &0.819 \\
\midrule[0.75pt]
Average &\textbf{0.863} &\textbf{0.681} &\textbf{0.957} &\textbf{0.851} &0.931 &0.841 \\
\bottomrule[1.5pt]

\end{tabular}
\vspace{5pt}
\caption{Agreement and Pearson correlation of \textsc{MMMG} evaluation with human annotations. ``IAA'' stands for inter-annotator agreement, ``agree'' stands for agreement and ``corr'' stands for Pearson correlation. We report Word Accuracy for text rendering, text editing and OCR tasks. Best results are in \textbf{bold}. \textsc{MMMG} achieves an average best human agreement of 0.944 with average inter-annotator agreement being 0.971. \textsc{GPT-4o} is the most human-aligned image evaluation model while $\mathrm{CLAPScore}_{\mathrm{audio}}$ is the most human-aligned audio evaluation method.}\label{tab:combined}

\end{table}
We report the agreement and Pearson correlation of \textsc{MMMG} with human annotation per task in Table \ref{tab:combined}. We exclude DreamSim and Whisper as they are widely recognized as established ``silver'' standards \citep{huang2024diffusion, mehrish2023review}.

\subsection{Full Benchmarking Results}
Evaluation results of 29 multimodal generation models on 55 tasks are listed in Table \ref{tab:imgae_all}, Table \ref{tab:imgae_text_all}, Table \ref{tab:audio_all} and Table \ref{tab:audio_text_all}, categorized by modalities.
We report the following additional findings:
\begin{itemize}[leftmargin=*]

\item Although image generation models generally maintain consistent rankings across various tasks, certain models exhibit notable weaknesses in specific areas. For instance, \textsc{Flux 1.1 Pro} performs particularly poorly when tasked with including unrelated objects in a scene, whereas \textsc{Imagen 3} struggles significantly with text rendering. These observations underscore the effectiveness of \textsc{MMMG} in pinpointing specific model weakness.

\item When comparing different interleaved image-text agent models, \textsc{Gemini 2.5} demonstrates superior planning capabilities over \textsc{GPT-4o}, resulting in a 51.8\% performance improvement with the image generator \textsc{GPT Image}.
\item Unified understanding-generation models such as \textsc{Janus} \citep{chen2025janus} are excluded from our interleaved image-text evaluation due to their requirement for manual modality selection, limiting their capability for automated, interleaved generation tasks. We also notice that models like \textsc{Anole} and \textsc{Seed-Llama} trained only on individual image generation and image understanding tasks can't follow instructions at all for interleaved image-text input. This highlight the importance of collecting more comprehensive image-text interleaved dataset for training.

\item Models are still consistently struggling with multimodal code understanding without image reference, with average model performance on Code (LateX) being only 0.059 and SVG being only 0.121. It is inspiring that agent-based models are able to understand multimodal code patches when presenting the original image as reference, where Gemini 2.5 + GPT Image achieves 0.383 accuracy. Being able to see the visualized multimodal codes is important for models to understand multimodal codes. This reveals a core difference in the multimodal code debugging code agent than the common code agent, where image signals (like screenshots) apart from unit tests should be included.

\item The natural speech-text interleaved model \textsc{Spirit LM} rarely scores above zero on evaluated tasks, suggesting it lacks adequate instruction tuning and consequently struggles to follow instructions effectively. Models like \textsc{GPT-4o-audio} and \textsc{Qwen2.5-Omni} \citep{xu2025qwen2} doesn't support customized speaker voice, thus can not be evaluated. Models like \textsc{Yue}, which are designed for text-to-song generation, may face challenges when are required to generate pure music.

\end{itemize}
\label{sec:full_res}
\begin{table}[!ht]\centering
\scriptsize
\tabcolsep=0.01cm
\begin{tabular}{lccccccccccccc}\toprule[1.5pt]
\textbf{Task}&\makecell{\textsc{Imagen}\\\textsc{3}} &\makecell{\textsc{Recraft}\\\textsc{v3}} &\makecell{\textsc{Luma}\\\textsc{Photon}} &\makecell{\textsc{Flux}\\\textsc{1.1 Pro}} &\makecell{\textsc{Ideo}\\\textsc{-gram 2}} &\textsc{Dalle 3} &\textsc{SD 3.5}   &\makecell{\textsc{Gemini 2}\\\textsc{Image}} &\makecell{\textsc{GPT}\\\textsc{Image}} &\textsc{BILP-o3}&\makecell{\textsc{Janus}\\\textsc{Pro}} &\makecell{\textsc{Gemini 2.5 }\\\textsc{Image}}\\\midrule[0.75pt]
Object Inclusion &0.838 &0.688 &0.831 &0.444 &0.863 &0.788 &0.544 &0.844 &0.869 &0.475 &0.706 & \textbf{0.919}\\
Object Exclusion &0.338 &0.300 &0.425 &0.325 &0.469 &0.244 &0.013 &0.281 &\textbf{0.819} &0.138 &0.063 & 0.681\\
Object Count &0.269 &0.319 &0.369 &0.375 &0.319 &0.119 &0.256 &0.356 &\textbf{0.569} &0.138 &0.263 & 0.450\\
Object Knowledge &0.494 &0.481 &0.656 &0.325 &0.419 &0.463 &0.150 &\textbf{0.706} &0.531 &0.238 &0.081 & 0.588\\
Object Commonsense &0.256 &0.306 &0.288 &0.206 &0.288 &0.288 &0.306 &0.275 &0.163 &0.194 &0.175 &\textbf{0.363}\\
Object Attribution &0.325 &0.206 &0.319 &0.256 &0.275 &0.294 &0.244 &0.375 &\textbf{0.619} &0.331 &0.363 & 0.475\\
Comparison Relation &0.588 &0.288 &0.488 &0.375 &0.475 &0.388 &0.150 &0.450 &0.600 &0.338 &0.200 & \textbf{0.650}\\
Universal Relation &0.425 &0.538 &0.638 &0.463 &0.500 &0.375 &0.350 &0.450 &\textbf{0.813} &0.238 &0.325 & 0.700\\
Relative Spatial&0.838 &0.625 &0.875 &0.663 &0.738 &0.550 &0.575 &0.750 &\textbf{0.988} &0.563 &0.588 & 0.900\\
Absolute Spatial&0.488 &0.388 &0.700 &0.488 &0.450 &0.225 &0.338 &\textbf{0.700} &0.675 &0.363 &0.563 & 0.588\\
Region Fill &0.484 &0.236 &0.628 &0.442 &0.375 &0.207 &0.320 &0.683 &0.762 &0.550 &0.415 & \textbf{0.788}\\
Border Fill &0.279 &0.353 &0.528 &0.349 &0.273 &0.350 &0.267 &0.450 &\textbf{0.651} &0.520 &0.363 & 0.368\\
Single TR &0.827 &0.994 &0.936 &0.901 &0.995 &0.661 &0.811 &0.997 &\textbf{1.000} &0.141 &0.544 & 0.990\\
Double TR &0.313 &0.422 &0.686 &0.528 &0.701 &0.215 &0.325 &0.745 &0.763 &0.008 &0.006 & \textbf{0.780}\\
Multi-lingual TR &0.351 &0.471 &0.440 &0.326 &0.483 &0.120 &0.330 &\textbf{0.817} &0.784 &0.000 &0.261 & 0.573\\
Average &0.474 &0.441 &0.587 &0.431 &0.508 &0.352 &0.332 &0.592 &\textbf{0.707} &0.282 &0.328 & 0.654\\
\bottomrule[1.5pt]
\end{tabular}
\vspace{5pt}
\caption{Benchmarking results of 10 models on 15 image generation tasks. Best results are in \textbf{bold}. \textsc{GPT-4o} significantly outperforms other image generation models.}\label{tab:imgae_all}

\end{table}
\begin{table}[!ht]\centering
\scriptsize
\tabcolsep=0.18cm
\begin{tabular}{lccccccc}\toprule[1.5pt]
\textbf{Task}&\makecell{\textsc{Seed}\\\textsc{Llama}} &\textsc{Anole} &\makecell{\textsc{GPT-4o} \\ \textsc{ + GPT Image}} &\makecell{\textsc{Gemini 2.5} \\ \textsc{ + GPT Image}} & \makecell{\textsc{Gemini 2} \\ \textsc{Image}}&\makecell{\textsc{GPT} \\ \textsc{Image}} &\makecell{\textsc{Gemini 2.5}\\\textsc{Image}}\\\midrule[0.75pt]
Semantic Consistency &0.000 &0.000 &0.613 &\textbf{0.763} &0.013 &0.675 &0.325\\
Multi-angle Consistency &0.000 &0.000 &0.230 &\textbf{0.461} &0.352 &0.448 &0.367\\
Multi-view Consistency &0.000 &0.000 &0.064 &\textbf{0.221} &0.143 &0.188 &0.191\\
Composition Consistency &0.000 &0.000 &\textbf{0.800} &0.738 &0.000 &0.075 &0.225\\
Decomposition Consistency &0.000 &0.000 &0.600 &\textbf{0.875} &0.013 &0.575 &0.638\\
Self Count &0.000 &0.038 &0.100 &\textbf{0.850} &0.213 &0.763 &0.000\\
Self Color Recognition &0.000 &0.000 &0.663 &0.700 &0.000 &\textbf{0.713} &0.088\\
Self Size Recognition &0.000 &0.000 &0.338 &0.600 &0.263 &\textbf{0.675} &0.463\\
Self Text Recognition &0.000 &0.000 &0.312 &\textbf{0.958} &0.101 &0.674 &0.569\\
Self Relative Spatial&0.000 &0.000 &0.538 &\textbf{0.725} &0.250 &0.425 &0.363\\
Self Absolute Spatial&0.000 &0.000 &0.475 &0.775 &0.100 &\textbf{0.825} &0.425\\
Text-image Order Control&0.150 &0.100 &0.913 &\textbf{0.925} &0.725 &0.500 &0.725\\
Interleaved Adding &0.154 &0.052 &0.394 &0.394 &0.545 &0.370 &\textbf{0.680}\\
interleaved Altering &0.179 &0.033 &0.566 &0.573 &0.609 &0.495 &\textbf{0.868}\\
Text Adding &0.000 &0.000 &0.097 &0.410 &0.444 &0.421 &\textbf{0.501}\\
Text Altering &0.046 &0.100 &0.077 &0.356 &0.182 &0.335 &\textbf{0.545}\\
Object Adding &0.165 &0.190 &0.470 &0.631 &0.748 &0.635 &\textbf{0.828}\\
Object Removing &0.350 &0.175 &0.415 &0.540 &0.605 &0.508 &\textbf{0.727}\\
Object Replacing &0.109 &0.121 &0.453 &0.627 &0.487 &0.650 &\textbf{0.722}\\
Object Altering &0.142 &0.000 &0.192 &0.352 &0.316 &0.360 &\textbf{0.503}\\
Interleaved Math &0.000 &0.000 &0.025 &\textbf{0.038} &0.000 &0.000 &0.000\\
Interleaved Code &0.000 &0.000 &0.071 &\textbf{0.224} &0.136 &0.202 &0.215\\
Average &0.059 &0.037 &0.382 &\textbf{0.579} &0.284 &0.478 &0.453\\
\bottomrule[1.5pt]
\end{tabular}
\vspace{5pt}
\caption{Benchmarking results of 6 models on 22 image-text interleaved generation tasks. Best results are in \textbf{bold}. Agent model \textsc{Gemini 2.5 Pro + GPT Image} is the best combination for consistent image sequence and coherent image-text pair generation. \textsc{Gemini 2.5 Image} as a modality-unified autoregressive model, performs best at image editing tasks.}\label{tab:imgae_text_all}

\end{table}
\begin{table}[!ht]\centering
\scriptsize
\tabcolsep=0.06cm
\begin{tabular}{lcccccccc}\toprule[1.5pt]
\textbf{Task}&\makecell{\textsc{Stable}\\\textsc{Audio}} &\makecell{\textsc{Audio}\\\textsc{LDM 2}} &\textsc{AudioGen} &\makecell{\textsc{Make-An}\\\textsc{-Audio 2}} &\textsc{Tango 2} &\textsc{MusicGen} &\makecell{\textsc{Tango}\\\textsc{Music}} &\textsc{Yue} \\\midrule[0.75pt]
Sound Begin-End &0.525 &0.450 &0.475 &\textbf{0.631} &0.525 &- &- &- \\
Sound Inclusion &\textbf{0.700} &0.413 &0.450 &0.575 &0.513 &- &- &- \\
Sound Reasoning &0.014 &0.014 &0.042 &\textbf{0.611} &0.194 &- &- &- \\
Sound Silence &0.063 &0.019 &0.019 &\textbf{0.131} &0.006 &- &- &- \\
Instrument Inclusion &0.817 &0.833 &- &- &- &0.833 &\textbf{0.950} &0.600 \\
Instrument Exclusion &0.225 &0.163 &- &- &- &0.200 &0.050 &\textbf{0.525} \\
Music Genre &0.488 &0.950 &- &- &- &0.625 &\textbf{0.925} &0.000 \\
Music Tempo &0.200 &0.010 &- &- &- &\textbf{0.620} &0.080 &0.040 \\
Music Intensity &\textbf{0.188} &0.013 &- &- &- &0.038 &0.075 &0.000 \\
Average &0.358 &0.318 &0.246 &\textbf{0.487} &0.310 &\textbf{0.463} &0.416 &0.233 \\
\bottomrule[1.5pt]
\end{tabular}
\vspace{5pt}
\caption{Benchmarking results of 8 models on 9 sound and music generation tasks. \textsc{Make-An-Audio 2} is the best audio generation model and the only model that can perform sound reasoning task; \textsc{MusicGen} is the best music generation model and the only model that can have tempo control.}\label{tab:audio_all}

\end{table}
\begin{table}[!ht]\centering
\scriptsize
\begin{tabular}{lccc}\toprule[1.5pt]
\textbf{Task} &\makecell{\textsc{Gemini 2.5}\\\textsc{+ VoxInstruct}} & \makecell{\textsc{Gemini 2.5}\\\textsc{+ VoiceLDM}} & \textsc{Spirit LM} \\\midrule[0.75pt]
Voice Attribution &\textbf{0.684} &0.568 &0.000 \\
Voice Replication &\textbf{0.625} &0.109 &0.002 \\
Speech Multi-lingual &\textbf{0.654} &- &- \\
Transcript Generation & \textbf{0.638}& 0.438& 0.200\\
Transcript Editing & 0.200& \textbf{0.350}& 0.000\\
Conversation Generation &\textbf{0.788} &0.375 &0.000 \\
Speech Translation & 0.155& \textbf{0.269}& 0.000\\
Speech Retrieval &0.188 &\textbf{0.421} &0.007 \\
Audio-Text Order Control &\textbf{0.520} &0.362 &0.023 \\
Average & \textbf{0.620} & 0.427 & 0.034 \\
\bottomrule[1.5pt]
\end{tabular}
\vspace{5pt}
\caption{Benchmarking results of 3 models on 9 speech-text interleaved generation tasks. Best results are in \textbf{bold}. Natural speech-text interleaved model \textsc{Spirit LM} does not have instruction following capability and get zero for most tasks. \textsc{VoxInstruct} is the best multi-functional speech synthesizer.}\label{tab:audio_text_all}

\end{table}
\begin{table}[!ht]\centering
\scriptsize
\tabcolsep=0.02cm
\begin{tabular}{lcccccccccccc}\toprule[1.5pt]
\textbf{Task}&\makecell{\textsc{Imagen}\\\textsc{3}} &\makecell{\textsc{Recraft}\\\textsc{v3}} &\makecell{\textsc{Luma}\\\textsc{Photon}} &\makecell{\textsc{Flux}\\\textsc{1.1 Pro}} &\makecell{\textsc{Ideo}\\\textsc{-gram 2}} &\textsc{Dalle 3} &\textsc{SD 3.5}   &\makecell{\textsc{Gemini 2}\\\textsc{Image}} &\makecell{\textsc{GPT}\\\textsc{Image}} &\textsc{BILP-o3}&\makecell{\textsc{Janus}\\\textsc{Pro}}&Average\\\midrule[0.75pt]
Object Inclusion &4.24 &3.16 &5.05 &5.43 &4.24 &4.24 &5.79 &3.08 &3.67 &4.90 &3.08 &4.26\\
Object Exclusion &4.24 &5.29 &7.21 &5.29 &6.44 &2.35 &2.45 &6.44 &1.22 &4.24 &1.41 &4.24\\
Object Count &3.08 &2.35 &4.18 &5.29 &5.05 &6.75 &1.23 &5.43 &\textbf{5.43} &5.10 &3.16 &4.28\\
Object Knowledge &1.23 &3.08 &2.35 &2.00 &3.67 &5.10 &7.21 &1.23 &5.05 &5.83 &2.35 &3.35\\
Object Commonsense &2.35 &6.44 &4.69 &2.35 &1.41 &4.69 &3.67 &2.00 &4.24 &4.64 &6.93 &3.95\\
Object Attribution &3.46 &3.08 &5.05 &5.79 &7.75 &1.22 &3.67 &4.47 &3.08 &2.35 &2.45 &3.85\\
Comparison Relation &2.45 &2.45 &\textbf{10.10} &6.33 &8.49 &8.37 &\textbf{6.93} &6.93 &4.00 &8.37 &8.00 &\textbf{6.85}\\
Universal Relation &2.83 &7.35 &6.17 &2.45 &\textbf{12.00} &\textbf{12.96} &5.66 &6.93 &2.45 &6.17 &\textbf{9.38} &6.76\\
Relative Spatial &4.69 &6.33 &6.33 &\textbf{10.87} &4.69 &6.93 &2.83 &6.93 &2.45 &10.10 &8.37 &6.41\\
Absolute Spatial &6.17 &4.69 &6.93 &4.69 &5.66 &\textbf{12.96} &2.45 &6.93 &4.90 &\textbf{10.87} &6.17 &6.58\\
Region Fill &4.56 &0.92 &5.26 &7.49 &5.31 &7.17 &5.06 &4.09 &3.28 &2.04 &4.34 &4.50\\
Border Fill &3.26 &7.54 &1.36 &8.39 &6.51 &5.52 &2.70 &\textbf{8.19} &4.07 &5.58 &1.62 &4.98\\
Single TR &5.36 &1.23 &6.13 &4.48 &0.98 &8.81 &7.46 &0.61 &0.00 &3.76 &5.66 &4.04\\
Double TR &2.49 &\textbf{11.66} &5.48 &9.87 &5.98 &6.20 &4.18 &1.91 &0.81 &0.30 &0.68 &4.50\\
Multi-lingual TR &\textbf{6.69} &2.52 &1.47 &1.98 &1.12 &2.82 &5.71 &3.94 &1.23 &0.00 &8.20 &3.24\\
Average &3.81 &4.54 &5.18 &5.51 &5.29 &\textbf{6.41} &4.47 &4.61 &3.06 &4.95 &4.79 &4.78\\

\bottomrule[1.5pt]
\end{tabular}
\vspace{5pt}
\caption{95\% relative confidence intervals of 9 models on 15 image generation tasks. The numbers are in percentile. Highest CIs are in \textbf{bold}.}\label{tab:imgae_ci}

\end{table}
\begin{table}[!ht]\centering
\scriptsize
\tabcolsep=0.2cm
\begin{tabular}{lccccccc}\toprule[1.5pt]
\textbf{Task}&\makecell{\textsc{Seed}\\\textsc{Llama}} &\textsc{Anole} &\makecell{\textsc{GPT-4o} \\ \textsc{ + GPT Image}} &\makecell{\textsc{Gemini 2.5} \\ \textsc{ + GPT Image}} & \makecell{\textsc{Gemini 2} \\ \textsc{Image}}&\makecell{\textsc{GPT} \\ \textsc{Image}}&Average\\\midrule[0.75pt]
Semantic Consistency &0.00 &0.00 &7.35 &2.45 &2.45 &6.33 &3.10 \\
Multi-angle Consistency &0.00 &0.00 &2.54 &0.55 &4.30 &2.69 &1.68 \\
Multi-view Consistency &0.00 &0.00 &1.03 &0.20 &2.01 &0.78 &0.67 \\
Composition Consistency &0.00 &0.00 &6.93 &\textbf{9.28} &0.00 &6.33 &3.76 \\
Decomposition Consistency &0.00 &0.00 &6.93 &2.83 &2.45 &6.33 &3.09 \\
Self Count &0.00 &2.45 &6.93 &6.93 &4.69 &8.37 &4.89 \\
Self Color Recognition &0.00 &0.00 &2.45 &5.66 &0.00 &4.69 &2.13 \\
Self Size Recognition &0.00 &0.00 &15.17 &6.93 &6.17 &6.33 &5.77 \\
Self Text Recognition &0.00 &0.00 &2.13 &0.56 &3.02 &\textbf{10.68} &2.73 \\
Self Relative Spatial&0.00 &0.00 &\textbf{16.19} &4.90 &4.00 &6.33 &\textbf{5.24} \\
Self Absolute Spatial&0.00 &0.00 &11.66 &2.83 &4.00 &4.90 &3.90 \\
Text-image Order Control&0.00 &\textbf{4.00} &4.69 &2.83 &6.33 &0.00 &2.97 \\
Interleaved Adding &0.08 &0.71 &0.75 &0.93 &2.56 &1.84 &1.14 \\
interleaved Altering &0.07 &1.04 &1.18 &1.60 &3.33 &1.29 &1.42 \\
Text Adding &0.00 &0.00 &1.57 &1.46 &0.91 &1.32 &0.88 \\
Text Altering &1.60 &0.00 &2.74 &0.70 &5.25 &3.40 &2.28 \\
Object Adding &1.38 &3.61 &7.25 &0.38 &3.50 &4.36 &3.41 \\
Object Removing &1.27 &3.44 &4.97 &1.40 &5.46 &2.42 &3.16 \\
Object Replacing &2.85 &3.46 &3.23 &4.00 &\textbf{9.85} &2.73 &4.35 \\
Object Altering &\textbf{5.85} &0.00 &4.50 &7.94 &1.23 &4.56 &4.01 \\
Interleaved Math &0.00 &0.00 &4.90 &2.45 &0.00 &0.00 &1.23 \\
Interleaved Code &0.00 &0.00 &3.22 &4.81 &3.23 &4.85 &2.68 \\
Average &0.60 &0.85 &\textbf{5.38} &3.25 &3.40 &4.11 & 2.93\\
\bottomrule[1.5pt]
\end{tabular}
\vspace{5pt}
\caption{95 relative confidence intervals of 6 models on 22 image-text interleaved generation tasks. The numbers are in percentile. Highest CIs are in \textbf{bold}.}\label{tab:imgae_text_ci}

\end{table}
\begin{table}[!ht]\centering
\scriptsize
\tabcolsep=0.15cm
\begin{tabular}{lccccccccc}\toprule[1.5pt]
\textbf{Task}&\makecell{\textsc{Stable}\\\textsc{Audio}} &\makecell{\textsc{Audio}\\\textsc{LDM 2}} &\textsc{AudioGen} &\makecell{\textsc{Make-An}\\\textsc{-Audio 2}} &\textsc{Tango 2} &\textsc{MusicGen} &\makecell{\textsc{Tango}\\\textsc{Music}} &\textsc{Yue} & Average\\\midrule[0.75pt]
Sound Begin-End &5.66 &2.83 &\textbf{13.12} &\textbf{8.95} &\textbf{5.25} &- &- &- &\textbf{7.16} \\
Sound Inclusion &\textbf{10.59} &\textbf{8.37} &5.66 &2.45 &5.13 &- &- &- &6.44 \\
Sound Reasoning &2.72 &2.72 &2.72 &5.44 &1.94 &- &- &- &3.11 \\
Sound Silence &2.45 &3.67 &1.23 &1.23 &0.63 &- &- &- &1.84 \\
Instrument Inclusion &6.26 &3.77 &- &0.00 &- &3.77 &3.27 &0.00 &2.84 \\
Instrument Exclusion &2.83 &7.35 &- &0.00 &- &5.66 &5.66 &\textbf{2.83} &4.05 \\
Music Genre &9.28 &5.66 &- &0.00 &- &\textbf{6.33} &\textbf{9.38} &0.00 &5.11 \\
Music Tempo &3.20 &1.96 &- &0.00 &- &5.06 &6.40 &0.00 &2.77 \\
Music Intensity &7.35 &2.45 &- &0.00 &- &4.69 &6.33 &0.00 &3.47 \\
Average &5.59 &4.31 &5.68 &2.01 &3.24 &5.10 &\textbf{6.21} &0.57 &4.09 \\
\bottomrule[1.5pt]
\end{tabular}
\vspace{5pt}
\caption{95 relative confidence intervals of 8 models on 9 sound and music generation tasks. The numbers are in percentile. Highest CIs are in \textbf{bold}.}\label{tab:audio_ci}

\end{table}
\begin{table}[!ht]\centering
\scriptsize
\begin{tabular}{lcccc}\toprule[1.5pt]
\textbf{Task} &\makecell{\textsc{Gemini 2.5}\\\textsc{+ VoxInstruct}} & \makecell{\textsc{Gemini 2.5}\\\textsc{+ VoiceLDM}} & \textsc{Spirit LM}  & Average\\\midrule[0.75pt]
Voice Attribution &4.14 &6.08 &0.01 &5.11 \\
Voice Replication &5.91 &2.80 &\textbf{0.13} &4.35 \\
Speech Multi-lingual &4.06 &0.17 &0.00 &2.12 \\
Transcript Generation &7.35 &\textbf{12.89} &0.00 &\textbf{10.12} \\
Transcript Editing &5.66 &12.00 &0.00 &8.83 \\
Conversation Generation &\textbf{7.35} &8.49 &0.00 &7.92 \\
Speech Translation &0.51 &1.16 &0.00 &0.84 \\
Speech Retrieval &3.37 &2.38 &0.00 &2.87 \\
Audio-Text Order Control &\textbf{6.93} &6.33 &0.00 &6.63 \\
Average &4.79 &5.74 &0.02 &5.27 \\
\bottomrule[1.5pt]
\end{tabular}
\vspace{5pt}
\caption{95 relative confidence intervals of 3 models on 9 speech-text interleaved generation tasks. The numbers are in percentile. Highest CIs are in \textbf{bold}.}\label{tab:audio_text_ci}

\end{table}

\subsection{Analysis}
\label{sec:analysis}
\paragraph{Task Separability}
We further analyze whether the tasks in \textsc{MMMG} are redundant from the perspective of model performance. As described in Section \ref{sec:mmmg:data}, our task templates are manually designed to test specific capabilities and are filtered to remove unrealistic, ambiguous, redundant, or low-application tasks. To study this issue more directly, we conduct a task separability analysis based on model accuracies.

We employ a standard approach in recent multi-task benchmark work \citep{ni2024mixeval1, zhang2024inherent}. For each \textsc{MMMG} task $t_i$, we form a vector $\mathbf{v}_i$ of per-model accuracies across all evaluated models (for image tasks, the 10 models in Table \ref{tab:imgae_all}). For each pair of tasks $(t_i, t_j)$ in the same modality, we compute the Pearson correlation between $\mathbf{v}_i$ and $\mathbf{v}_j$. Intuitively, if two tasks always rank models in almost the same way, they behave like a single task; if correlations are lower, they capture complementary signals. Using the nine image generation models shared with Table~3, we first compute cross-benchmark Spearman correlations between GenEval, DrawBench, and GenAI-Bench: $\rho(\text{GenEval}, \text{GenAI-Bench}) = 0.778$, $\rho(\text{DrawBench}, \text{GenAI-Bench}) = 0.754$, $\rho(\text{GenEval}, \text{DrawBench}) = 0.569$.

We also compute the correlation of all image task pairs in \textsc{MMMG} as shown in Table \ref{tab:task_separability}. Across all such task pairs, we find that only 38 pairs (18.01\% of all pairs) have correlation higher than 0.778. In other words, more than 80\% of \textsc{MMMG} task pairs are less correlated with each other than two popular image benchmarks are with each other. This indicates that our tasks are not a collection of near-duplicate prompts: they provide more diverse behavioral signals than entire existing benchmarks do. Since there are no baseline benchmarks for reference in other modalities, we only present the correlation between image tasks.

\begin{table}[!ht]
\centering
\scriptsize
\tabcolsep=0.1cm
\begin{tabular}{lccccccccccccccc}
\toprule[1.5pt]
\textbf{Task} & \makecell{\textbf{Obj.}\\\textbf{Inc.}} & \makecell{\textbf{Obj.}\\\textbf{Exc.}} & \makecell{\textbf{Obj.}\\\textbf{Cou.}} & \makecell{\textbf{Obj.}\\\textbf{Kno.}} & \makecell{\textbf{Obj.}\\\textbf{Com.}} & \makecell{\textbf{Obj.}\\\textbf{Att.}} & \makecell{\textbf{Com.}\\\textbf{Rel.}} & \makecell{\textbf{Uni.}\\\textbf{Rel.}} & \makecell{\textbf{Rel.}\\\textbf{Spa.}} & \makecell{\textbf{Abs.}\\\textbf{Spa.}} & \makecell{\textbf{Reg.}\\\textbf{Fill}} & \makecell{\textbf{Bor.}\\\textbf{Fill}} & \makecell{\textbf{Sin.}\\\textbf{TR}} & \makecell{\textbf{Dou.}\\\textbf{TR}} & \makecell{\textbf{Mul.}\\\textbf{TR}} \\

\midrule[0.75pt]
Object Include      & -      & 0.677 & 0.464 & 0.712 & 0.441 & 0.557 & 0.734 & 0.654 & 0.720 & 0.501 & 0.473 & 0.096 & 0.570 & 0.611 & 0.627 \\
Object Exclude      & 0.677  & -     & 0.859 & 0.671 & 0.232 & 0.742 & 0.884 & 0.930 & 0.889 & 0.540 & 0.691 & 0.389 & 0.601 & 0.799 & 0.656 \\
Object Count        & 0.464  & 0.859 & -     & 0.609 & 0.198 & 0.623 & 0.673 & 0.919 & 0.857 & 0.680 & 0.696 & 0.378 & 0.767 & 0.894 & 0.838 \\
Object Knowl.       & 0.712  & 0.671 & 0.609 & -     & 0.466 & 0.363 & 0.764 & 0.671 & 0.700 & 0.529 & 0.530 & 0.309 & 0.634 & 0.772 & 0.663 \\
Object Common.      & 0.441  & \textbf{0.232} & \textbf{0.198} & 0.466 & -      & \textbf{-0.120} & 0.304 & \textbf{0.298} & \textbf{0.196} & \textbf{-0.068} & \textbf{0.087} & \textbf{-0.458} & 0.477 & 0.453 & \textbf{0.220} \\
Object Attribute    & 0.557  & 0.742 & 0.623 & 0.363 & -0.120 & -     & 0.664 & 0.664 & 0.740 & 0.615 & 0.823 & 0.612 & 0.188 & 0.449 & 0.550 \\
Compare Relation    & 0.734  & 0.884 & 0.673 & 0.764 & 0.304 & 0.664 & -     & 0.735 & 0.879 & 0.518 & 0.729 & 0.264 & 0.465 & 0.695 & 0.525 \\
Universal Relation  & 0.654  & 0.930 & 0.919 & 0.671 & 0.298 & 0.664 & 0.735 & -     & 0.881 & 0.621 & 0.637 & 0.395 & 0.744 & 0.844 & 0.750 \\
Relative Spatial    & 0.720  & 0.889 & 0.857 & 0.700 & 0.196 & 0.740 & 0.879 & 0.881 & -     & 0.763 & 0.807 & 0.400 & 0.619 & 0.789 & 0.729 \\
Absolute Spatial    & 0.501  & 0.540 & 0.680 & 0.529 & -0.068 & 0.615 & 0.518 & 0.621 & 0.763 & -     & 0.802 & 0.515 & 0.468 & 0.633 & 0.745 \\
Region Fill         & 0.473  & 0.691 & 0.696 & 0.530 & 0.087 & 0.823 & 0.729 & 0.637 & 0.807 & 0.802 & -     & 0.557 & 0.252 & 0.611 & 0.593 \\
Border Fill         & \textbf{0.096} & 0.389 & 0.378 & \textbf{0.309} & \textbf{-0.458} & 0.612 & \textbf{0.264} & 0.395 & 0.400 & 0.515 & 0.557 & -     & \textbf{-0.113} & \textbf{0.201} & 0.281 \\
Single TR           & 0.570  & 0.601 & 0.767 & 0.634 & 0.477 & 0.188 & 0.465 & 0.744 & 0.619 & 0.468 & 0.252 & -0.113 & -     & 0.859 & 0.817 \\
Double TR           & 0.611  & 0.799 & 0.894 & 0.772 & 0.453 & 0.449 & 0.695 & 0.844 & 0.789 & 0.633 & 0.611 & 0.201 & 0.859 & -     & 0.851 \\
Multi-lingual TR    & 0.627  & 0.656 & 0.838 & 0.663 & 0.220 & 0.550 & 0.525 & 0.750 & 0.729 & 0.745 & 0.593 & 0.281 & 0.817 & 0.851 & - \\
\bottomrule[1.5pt]
\end{tabular}
\caption{Image generation task separability analysis based on model performance Pearson correlations. The lowest correlation are in \textbf{bold}.}
\label{tab:task_separability}
\end{table}

% different experiment setup
\paragraph{Interleaved System Prompt.} To investigate whether autoregressive models' capabilities in generating the desired number and order of modalities can be improved, we conducted experiments with \textsc{Gemini 2 Image} using the planning system prompt detailed in Table \ref{tab:gemini_prompt}. The experimental results, summarized in Table \ref{tab:gemini_res}, indicate that incorporating system prompts emphasizing modality count and order does not consistently lead to positive outcomes. Generally, adding a system prompt negatively impacts image generation quality, as the models shift their focus away from optimizing visual quality. Conversely, image editing tasks benefit from the addition of system prompts since without such prompts, models frequently generate multiple images unnecessarily. Nonetheless, system prompts do not effectively support generating sequential images or integrated image-text pairs, because models continue to intermix multiple images during generation, as illustrated in Figure \ref{fig:failure}.

\begin{table}[!ht]
\centering
% \scriptsize
\begin{tabular}{p{0.95\linewidth}}
\toprule[1.5pt]

You are a multimodal assistant capable of generating interleaved text and images based on user instructions.

\begin{itemize}[leftmargin=*]
    \item Follow the required modality structure and number in user’s instruction exactly, especially when multiple images are implied or requested.
    \item Generate separate images for each described part, do not combine multiple concepts into one image unless told to.
    \item Interleave images and text in the order described.
\end{itemize}

Your goal is to match the user’s intent with exact number and sequence of image and text.\\
\bottomrule[1.5pt]
\end{tabular}
\vspace{5pt}
\caption{System prompt used to make \textsc{Gemini Image} output correct modality order and number.} \label{tab:gemini_prompt}
\end{table}`
\begin{table}[!ht]\centering
\scriptsize
\tabcolsep=0.03cm
\begin{minipage}{0.48\textwidth}
\centering
\begin{tabular}{lcc}\toprule[1.5pt]
\textbf{Task} & \makecell{\textsc{Gemini Image}\\w/ prompt} & \makecell{\textsc{Gemini Image}\\w/o prompt} \\
\midrule[0.75pt]
Semantic Consistency &\textbf{0.263} &0.013 \\
Multi-Angel Consistency &0.135 &\textbf{0.352} \\
Multi-View Consistency &0.094 &\textbf{0.143} \\
Compose Consistency &\textbf{0.013} &0.000 \\
Decompose Consistency &0.000 &\textbf{0.013} \\
Interleaved Object Adding &0.399 &\textbf{0.545} \\
Interleaved Color Modifying &0.486 &\textbf{0.609} \\
Text Editing &\textbf{0.423} &0.283 \\
Object Adding &0.622 &\textbf{0.748} \\
Object Removing &0.485 &\textbf{0.605} \\
Object Modifying &0.468 &\textbf{0.487} \\
Self Count &\textbf{0.275} &0.213 \\
Self Color &\textbf{0.113} &0.000 \\
Self Size &0.188 &\textbf{0.263} \\
Self OCR &\textbf{0.335} &0.101 \\
Self Relative Spatial &0.138 &\textbf{0.250} \\
Self Absolute Spatial &\textbf{0.175} &0.100 \\
Interleaved Math &\textbf{0.000} &\textbf{0.000} \\
Interleaved Code &0.110 &\textbf{0.136} \\
Image-Text Order &\textbf{0.725} &\textbf{0.725}\\
Average &0.273 &\textbf{0.279} \\
\bottomrule[1.5pt]
\end{tabular}
\end{minipage}\hfill
\begin{minipage}{0.48\textwidth}
\centering
\tabcolsep=0.03cm
\begin{tabular}{lcc}\toprule[1.5pt]
\textbf{Task} & \makecell{\textsc{Gemini Image}\\w/ prompt} & \makecell{\textsc{Gemini Image}\\w/o prompt} \\
\midrule[0.75pt]
Object Inclusion &\textbf{0.888} &0.875 \\
Object Exclusion &\textbf{0.400} &0.313 \\
Object Count &\textbf{0.500} &0.450 \\
Object Reasoning &\textbf{0.813} &0.825 \\
Object Attribution &\textbf{0.475} &\textbf{0.475} \\
Comparison Relation &\textbf{0.475} &0.450 \\
Universal Relation &\textbf{0.488} &0.450 \\
Relative Spacial Relation &\textbf{0.850} &0.750 \\
Absolute Spacial Relation &\textbf{0.738} &0.700 \\
Region Fill &0.585 &\textbf{0.683} \\
Border Fill &\textbf{0.459} &0.450 \\
Single Text Rendering &0.945 &\textbf{0.997} \\
Double Text Rendering &\textbf{0.800} &0.745 \\
Multi-lingual Text Rendering &0.691 &\textbf{0.817} \\
Average &\textbf{0.650} &0.641 \\
\bottomrule[1.5pt]
\end{tabular}
\end{minipage}
\vspace{5pt}
\caption{Comparison of \textsc{Gemini Image} performance with and without system prompt on image generation (right) and interleaved image-text generation (left) tasks. Best results are in \textbf{bold}. System prompt does not always have positive impact.}
\label{tab:gemini_res}
\end{table}

\paragraph{Variance Control.} 
\label{sec:variance}
To validate the evaluation robustness of \textsc{MMMG}, we present the 95\% confidence intervals for each task in Table \ref{tab:imgae_ci}, Table \ref{tab:imgae_text_ci}, Table \ref{tab:audio_ci} and Table \ref{tab:audio_text_ci}. A sample size of 4 can substantially reduce variance, with the average relative confidence interval of all tasks and models being 4.05\%. While some individual models show higher variance on certain tasks, this reflects the inherent robustness differences of the models themselves rather than evaluation instability. Importantly, when averaged across all models, the maximum 95\% CI is only 10.12\% of all tasks and 6.41\% of all models. The, demonstrating that \textsc{MMMG} provides statistically robust evaluation across the full spectrum of capabilities.

\subsection{Evaluation Failure Analysis}
Apart from the numbers we reported in Table \ref{tab:combined}, we also look at qualitative cases where our evaluation models fail. We present error analyses of CLAPScore and \textsc{WavLM} in Tables \ref{tab:clap_error_analysis} and \ref{tab:wavlm_error_analysis}. Results point out that improving accuracy on synthetic audio and increase the diversity of reference audio set is the key direction for future evaluation model improvement. 
\begin{table}[!ht]
\centering
\small
\renewcommand{\arraystretch}{1.2}
\begin{tabular}{p{0.15\textwidth} p{0.12\textwidth} p{0.3\textwidth} p{0.3\textwidth}}
\toprule[1.5pt]
\textbf{Error Category} & \textbf{\% in All} & \textbf{Observation} & \textbf{Plausible Causes} \\
\midrule[0.75pt]
Audio generation failure & 20.0\% & Generated audio consists of pure noise with no meaningful structure. & CLAPScore is not OOD-generalizable. When generation quality is low, it cannot make reliable judgments. \\

Audio generation quality unsatisfactory & 26.7\% & Generated audio contains recognizable and reasonable sounds but is overall of low quality with distortion. For example, the models repeatedly failed to generate proper reggae music, often producing tracks dominated by drums. & CLAPScore is not OOD-generalizable. When generation quality is low, it cannot make reliable judgments. \\

Lack of fine-grained understanding & 26.7\% & Some generated audio includes fine-grained details that are easily recognizable by humans, such as brief dog barks or brief sneezes, but their duration is too short. Others contain noisy background such as sirens in loud traffic, which is treated as a meaningless piece by the CLAP model. & Since CLAPScore computes the similarity of the whole generated audio with the reference audio, short duration or noisy background will underestimate such similarity. Future methods could attempt to locate the target sound first before computing similarity. \\

Underrepresented reference audio & 26.7\% & Some generated audio is not representative in the reference audio. A peaceful drum sequence is not considered as a drum piece by the CLAP model since all the reference drum music is metallic-like. & The reference datasets we use are standard and may not fully capture a sound, an instrument, or a music genre. Enriching the dataset diversity is important for more reliable evaluation. \\
\bottomrule[1.5pt]
\end{tabular}
\caption{Error analysis of CLAPScore. The total error rate is 7.7\%.}
\label{tab:clap_error_analysis}
\end{table}
\begin{table}[!ht]
\centering
\small
\renewcommand{\arraystretch}{1.2}
\begin{tabular}{p{0.15\textwidth} p{0.12\textwidth} p{0.3\textwidth} p{0.3\textwidth}}
\toprule[1.5pt]
\textbf{Error Category} & \textbf{\% in All} & \textbf{Observation} & \textbf{Plausible Causes} \\
\midrule[0.75pt]
Generated speech with noise or distortion & 21.4\% & Some generated speeches contain noticeable background white noise. Sometimes, the noise is intermittent volume fluctuation (sometimes slightly lower, sometimes higher) or long pauses, which may interfere with the model's judgment. & \textsc{WavLM} is not OOD-generalizable. When generation quality is low, it cannot make reliable judgments. \\

Generated speech with synthetic pattern & 57.1\% & Some generated speeches can be considered as coming from the same speaker, but with clear synthetic patterns, making it hard even for human judges to decide whether the generated speeches are from the same speaker. & \textsc{WavLM} is not OOD-generalizable. It is trained on real-world human voice and may not effectively judge synthetic speeches. \\

Computed similarity score close to threshold & 21.4\% & The generated speeches are very similar to the references, aside from slight white noise. This likely caused the model's judgment to be only minimally affected, placing the scores near the decision threshold. & \textsc{WavLM} is not well-calibrated. It will only give high scores to the same speaker and low scores to different speakers. When the score is close to the threshold, it is likely to make mistakes. \\
\bottomrule[1.5pt]
\end{tabular}
\caption{Error analysis of \textsc{WavLM}. The total error rate is 13.7\%.}
\label{tab:wavlm_error_analysis}
\end{table}

\subsection{Generation Model Failure Analysis}
We add a dedicated analysis of multimodal generation model failures, including (1) an explicit error taxonomy, (2) error-type distributions and statistics, and (3) analysis of likely causes behind these differences. We investigate 246 image generation errors and 330 interleaved image-text generation errors on 4 models. The image generation error distributions are summarized in Tables~\ref{tab:image_failure_analysis_1}--\ref{tab:image_failure_analysis_3}, and the interleaved image-text generation error distributions are summarized in Tables~\ref{tab:interleaved_failure_analysis_1}--\ref{tab:interleaved_failure_analysis_4}.

For image generation, \textsc{Gemini 2.0} and \textsc{GPT Image} show similar error patterns. Most failures are rooted in insufficient training data and training data bias, including under-training, coarse captions, distributional bias, and spurious correlations. This suggests that architecture advantages mainly come from scalability, i.e., the ability to effectively consume larger amounts of data \citep{tianvisual}.

For interleaved image-text generation, the two architectures exhibit distinct failure patterns. ARMs mainly suffer from poor sequential generation control: they frequently generate the wrong number of images (13.9\%) and fail to maintain temporal consistency (8.4\%), suggesting difficulty in tracking state and count during long-sequence generation. They also have poor output format control, generating text in the required format (8.4\%) and sometimes combining multiple requested images into a single entangled output (1.3\%), indicating weak control over structured outputs. In contrast, agent-based models mainly suffer from poor cross-image consistency and poor global editing control. Because they rely on independent image generation tools, they often produce inconsistent image sequences (8.7\%) or images that contradict their own relation descriptions in text (15.2\%), reflecting the lack of cross-image state maintenance and self-correction mechanisms. In image editing, they are also more prone to unintended global changes outside the target region (7.6\%), suggesting that the planning model cannot effectively constrain the output scope of the generation tool.

\begin{table*}[!ht]
\centering
\small
\tabcolsep=0.08cm
\renewcommand{\arraystretch}{1.2}
\begin{tabular}{p{0.15\textwidth} p{0.27\textwidth} p{0.27\textwidth} p{0.08\textwidth} p{0.08\textwidth} p{0.08\textwidth}}
\toprule[1.5pt]
\makecell{\textbf{Error}\\\textbf{Category}} & \makecell{\textbf{Examples}} & \makecell{\textbf{Possible}\\\textbf{Reasons}} & \makecell{\textbf{\# (\%)}\\\textbf{of All}\\\textbf{Err.}} & \makecell{\textbf{\# (\%)}\\\textbf{of Gem}\\\textbf{Err.}} & \makecell{\textbf{\# (\%)}\\\textbf{of GPT}\\\textbf{Err.}} \\
\midrule[0.75pt]
Fail to render required scene properly & Instruction: ``Create an image of a waterfall flowing into the sky, containing one pumpkin and one suitcase.'' Error: The waterfall is rendered normally, not flowing into the sky. & The model relies heavily on physical laws and realistic priors embedded in the training data, and struggles to synthesize scenes that violate common-sense physics. & \errstat{3}{1.2\%} & \errstat{0}{0.0\%} & \errstat{3}{3.0\%} \\

Fail to render objects of the required number & Instruction: ``Generate an image of a historical site. Please include a single spaceship and a giraffe in the image.'' Error: Two spaceships are generated even though the instruction asks for one. & The model demonstrates limited numeracy and counting capabilities. This may be due to the attention mechanism, which often fails to disentangle identical object instances and leads to superfluous objects. & \errstat{49}{19.9\%} & \errstat{27}{18.5\%} & \errstat{22}{22.0\%} \\

Object rendered with distortion or unrealistic appearance & Instruction: ``Create an image of an endless mirrored hallway, with one cactus and one saxophone.'' Error: The mirrored hallway is distorted. & This may be due to under-training. Models may not be trained on the target objects enough to reproduce full details. & \errstat{15}{6.1\%} & \errstat{6}{4.1\%} & \errstat{9}{9.0\%} \\

Required object not present, partially present, or wrong object present & Instruction: ``Create an image of a whale flying through a sunset sky, featuring one suitcase and one mailbox.'' Error: The image does not include a suitcase. & Failed instruction following may indicate under post-training. Instead of generating the required image, the model generates the most plausible image. & \errstat{1}{0.4\%} & \errstat{0}{0.0\%} & \errstat{1}{1.0\%} \\

Unable to remove the correlated object & Instruction: ``Generate an image of people camping. Do not include tents in the image.'' Error: The generated image still contains tents. & Negative constraints paradoxically activate associated concepts in the latent space. In addition, strong semantic co-occurrence between scenes and objects (spurious correlation) makes it difficult to decouple these associations. & \errstat{31}{12.6\%} & \errstat{24}{16.4\%} & \errstat{7}{7.0\%} \\

Fail to render required attribute (single object) & Instruction: ``Generate an image of a single chair with 5 legs.'' Error: The generated chair has 4 legs. & Strong visual priors, such as the normative four-legged chair, override textual prompts and create a conflict between internal knowledge and counterfactual instructions. & \errstat{28}{11.4\%} & \errstat{20}{13.7\%} & \errstat{8}{8.0\%} \\

Fail to bind object attributes correctly (multiple objects) & Instruction: ``Orange elephant with purple polka dots.'' Error: The colors are mixed on the elephant's body. & Features from one entity, such as color or texture, erroneously propagate to adjacent objects. This is likely due to coarse captions in training data, which omit detailed object attribution. & \errstat{17}{6.9\%} & \errstat{9}{6.2\%} & \errstat{8}{8.0\%} \\
\bottomrule[1.5pt]
\end{tabular}
\caption{Generation model failure analysis on image generation tasks (Part I). Analyzed models are \textsc{Gemini Image 2.0} and \textsc{GPT Image}.}
\label{tab:image_failure_analysis_1}
\end{table*}
\begin{table*}[!ht]
\centering
\small
\tabcolsep=0.08cm
\renewcommand{\arraystretch}{1.2}
\begin{tabular}{p{0.15\textwidth} p{0.27\textwidth} p{0.27\textwidth} p{0.08\textwidth} p{0.08\textwidth} p{0.08\textwidth}}
\toprule[1.5pt]
\makecell{\textbf{Error}\\\textbf{Category}} & \makecell{\textbf{Examples}} & \makecell{\textbf{Possible}\\\textbf{Reasons}} & \makecell{\textbf{\# (\%)}\\\textbf{of All}\\\textbf{Err.}} & \makecell{\textbf{\# (\%)}\\\textbf{of Gem}\\\textbf{Err.}} & \makecell{\textbf{\# (\%)}\\\textbf{of GPT}\\\textbf{Err.}} \\
\midrule[0.75pt]
Fail to apply attribute requirements to all objects (multiple objects) & Instruction: ``Generate an image of birds on a wire where all birds are facing the same direction.'' Error: Not all birds are facing the same direction. & The model has difficulty maintaining attribute consistency across multiple entities. Stochastic generation processes often fail to enforce constraints uniformly on every instance. & \errstat{6}{2.4\%} & \errstat{4}{2.7\%} & \errstat{2}{2.0\%} \\

Fail to apply attribute requirements only to the required objects (multiple objects) & Instruction: ``Generate an image of a bookshelf where all books are standing vertically except one lying horizontally.'' Error: More than one book is lying horizontally. & High logical complexity hampers the execution of exception logic, causing the model to over-generalize rules to excluded entities. & \errstat{6}{2.4\%} & \errstat{5}{3.4\%} & \errstat{1}{1.0\%} \\

Wrong / insufficient knowledge to reason the target object, wrong object generated & Instruction: ``Create an image featuring only the national flag of a country that has a major city located on both the European and Asian continents.'' Error: The model generates the French flag instead of the Turkish flag. & Unlike LLMs, image generation models are usually not trained on extensive world knowledge, so they can only follow the literal instruction and may fail to infer the intended target object. & \errstat{20}{8.1\%} & \errstat{8}{5.5\%} & \errstat{12}{12.0\%} \\

Wrong / insufficient knowledge to reason the target object, multiple objects generated & Instruction: ``Generate an image of a small, handheld stringed instrument, commonly used in folk and country music and important in black American music.'' Error: The model generates multiple instruments. & This shows signs of reward hacking in post-training. When the reward model does not punish guessing behavior, the model may generate multiple objects instead of following the instruction. & \errstat{1}{0.4\%} & \errstat{0}{0.0\%} & \errstat{1}{1.0\%} \\

Fail to render the required logical relation between objects & Instruction: ``Create an image featuring a single nail and a single snake, and the nail is longer than the snake.'' Error: The nail is still shorter than the snake. & The unrealistic logical relation in the prompt conflicts with the model's internal semantic priors, and the model favors the prior. & \errstat{25}{10.2\%} & \errstat{17}{11.6\%} & \errstat{8}{8.0\%} \\

Fail to render the required spatial relation between objects & Instruction: ``Generate an image of a peaceful garden, with a single watering can at the upper right quarter of the image.'' Error: The watering can is only on the right side. & Models are trained on image-caption pairs that stress semantic consistency and overlook perception accuracy. & \errstat{12}{4.9\%} & \errstat{6}{4.1\%} & \errstat{6}{6.0\%} \\

Fail to render accurate image format & Instruction: ``Generate a waterfall in a lush rainforest. The entire image should be surrounded by a simple and flat, solid and cyan border of approximately 10\% of the image width on all sides.'' Error: The model generates a border of about 30\%. & The continuous nature of image generation makes precise geometric or color thresholds difficult to enforce. Models tend to prioritize semantic coherence over strict low-level geometric constraints. & \errstat{9}{3.7\%} & \errstat{7}{4.8\%} & \errstat{2}{2.0\%} \\
\bottomrule[1.5pt]
\end{tabular}
\caption{Generation model failure analysis on image generation tasks (Part II). Analyzed models are \textsc{Gemini Image 2.0} and \textsc{GPT Image}.}
\label{tab:image_failure_analysis_2}
\end{table*}
\begin{table*}[!ht]
\centering
\small
\tabcolsep=0.08cm
\renewcommand{\arraystretch}{1.2}
\begin{tabular}{p{0.15\textwidth} p{0.27\textwidth} p{0.27\textwidth} p{0.08\textwidth} p{0.08\textwidth} p{0.08\textwidth}}
\toprule[1.5pt]
\makecell{\textbf{Error}\\\textbf{Category}} & \makecell{\textbf{Examples}} & \makecell{\textbf{Possible}\\\textbf{Reasons}} & \makecell{\textbf{\# (\%)}\\\textbf{of All}\\\textbf{Err.}} & \makecell{\textbf{\# (\%)}\\\textbf{of Gem}\\\textbf{Err.}} & \makecell{\textbf{\# (\%)}\\\textbf{of GPT}\\\textbf{Err.}} \\
\midrule[0.75pt]
Fail to render image format at all & Instruction: ``Generate a space scene with planets and nebulae. The entire image should be surrounded by a simple and flat, solid and black border of approximately 10\% of the image width on all sides.'' Error: The scene is generated without any border. & Formatting constraints are ignored because the model interprets them as soft suggestions rather than hard rules. This is likely an OOD problem, where such image formats are rarely requested in training data. & \errstat{1}{0.4\%} & \errstat{0}{0.0\%} & \errstat{1}{1.0\%} \\

Rendered text with misspelling & Instruction: ``Generate the text `always move forward'.'' Error: The generated text is misspelled as ``always move forwad.'' & Text rendering is a known problem for image generation. Models operate at the pixel level rather than the character level, and without explicit orthographic verification they tend to generate pseudo-text that is visually similar but orthographically incorrect. & \errstat{10}{4.1\%} & \errstat{6}{4.1\%} & \errstat{4}{4.0\%} \\

Rendered text with duplication & Instruction: ``Generate the text `capture the moment'.'' Error: The generated text becomes ``capture the the moment.'' & Repetition artifacts arise from loops or redundancies in the text-generation attention mechanism. & \errstat{3}{1.2\%} & \errstat{2}{1.4\%} & \errstat{1}{1.0\%} \\

Required text not rendered & Instruction: ``Generate the text `Washington WA 98105 Evergreen State'.'' Error: No text is generated. & Lack of text rendering training makes the model neglect text requirements. & \errstat{0}{0.0\%} & \errstat{0}{0.0\%} & \errstat{0}{0.0\%} \\

Required text rendered in distortion & Instruction: ``Generate the text `Enjoy The Little Things Always'.'' Error: The word ``Enjoy'' has distorted glyphs. & Lack of text rendering training makes the model unable to generate high-quality text. & \errstat{1}{0.4\%} & \errstat{1}{0.7\%} & \errstat{0}{0.0\%} \\

Required text in the wrong place or overflow & Instruction: ``Generate an image of exactly two posters on an office wall...'' Error: The model generates three posters, and the second text appears on the middle poster. & Lack of text rendering training makes the model unable to place text correctly. Adding a layout controller or training on such data may help. & \errstat{5}{2.0\%} & \errstat{2}{1.4\%} & \errstat{3}{3.0\%} \\

Required text in wrong language & Instruction: ``Generate an image of a notebook and the only text on it is `\begin{CJK*}{UTF8}{gbsn}不要忘记\end{CJK*}'.'' Error: The model generates symbol-like Chinese characters that are unreadable. & Distributional bias in training data and visual similarity between character sets can lead to language performance bias during generation. & \errstat{3}{1.2\%} & \errstat{2}{1.4\%} & \errstat{1}{1.0\%} \\
\bottomrule[1.5pt]
\end{tabular}
\caption{Generation model failure analysis on image generation tasks (Part III). Analyzed models are \textsc{Gemini Image 2.0} and \textsc{GPT Image}.}
\label{tab:image_failure_analysis_3}
\end{table*}
\begin{table*}[!ht]
\centering
\small
\tabcolsep=0.08cm
\renewcommand{\arraystretch}{1.2}
\begin{tabular}{p{0.15\textwidth} p{0.27\textwidth} p{0.27\textwidth} p{0.08\textwidth} p{0.08\textwidth} p{0.08\textwidth}}
\toprule[1.5pt]
\makecell{\textbf{Error}\\\textbf{Category}} & \makecell{\textbf{Examples}} & \makecell{\textbf{Possible}\\\textbf{Reasons}} & \makecell{\textbf{\# (\%)}\\\textbf{of All}\\\textbf{Err.}} & \makecell{\textbf{\# (\%)}\\\textbf{of Gem}\\\textbf{Err.}} & \makecell{\textbf{\# (\%)}\\\textbf{of Age}\\\textbf{Err.}} \\
\midrule[0.75pt]
Generate wrong number of images & Instruction: ``Using the provided image as the reference angle, create four additional images...'' Error: The model generates only three images and misses the final view. & Difficulty in precise counting and planning; the model loses track of the count during the sequential generation loop. & \errstat{40}{12.2\%} & \errstat{33}{13.9\%} & \errstat{7}{7.6\%} \\

Only image description provided but no image & Instruction: ``Create an image depicting a musician's room that includes exactly one microphone, one guitar case, and one music stand...'' Error: The model provides the text evaluation but does not generate the actual image. & Tool-use failure; the model prioritizes text reasoning instead of actually calling the image generation tool. & \errstat{2}{0.6\%} & \errstat{2}{0.8\%} & \errstat{0}{0.0\%} \\

Low image quality or distorted, unrealistic object & Instruction: ``Based on the provided image showing a frontal view, create four more images depicting the scene from specific angles...'' Error: Later views become severely distorted and blurry. & This may be due to under-training. Models may not be trained on the target objects enough to reproduce full details. & \errstat{4}{1.2\%} & \errstat{4}{1.7\%} & \errstat{0}{0.0\%} \\

Combine multiple images together instead of generating multiple images & Instruction: ``Based on the provided image showing the frontal view, create four additional images...'' Error: The model generates a single image grid instead of four separate images. & Models, especially modality-unified ARMs, are likely to entangle multiple images in one output due to continuous latent representations and a training bias toward single-turn unified outputs. & \errstat{3}{0.9\%} & \errstat{3}{1.3\%} & \errstat{0}{0.0\%} \\

Inconsistent image sequence: objects, style, or scene change during generation & Instruction: ``Create four images that sequentially show the addition of a passport, a map, a camera, and a pair of sunglasses...'' Error: Previously added objects disappear and the suitcase style changes. & Agent models do not show strong planning capabilities on image generation tasks. ARMs also lack cross-image consistency mechanisms, so state is not maintained across steps. & \errstat{16}{4.9\%} & \errstat{8}{3.4\%} & \errstat{8}{8.7\%} \\

Image sequence shows no variance: image repetition & Instruction: ``Using the provided image as the reference angle, create four additional images...'' Error: All generated images are identical to the original reference view. & Mode output collapse. This is likely related to the same phenomenon as repetitive outputs in LLMs, and may indicate lack of post-training. & \errstat{11}{3.3\%} & \errstat{8}{3.4\%} & \errstat{3}{3.3\%} \\
\bottomrule[1.5pt]
\end{tabular}
\caption{Generation model failure analysis on interleaved image-text generation tasks (Part I). Analyzed models are \textsc{Gemini Image 2.0} and \textsc{Gemini 2.5 Pro} + \textsc{GPT Image} Agent.}
\label{tab:interleaved_failure_analysis_1}
\end{table*}
\begin{table*}[!ht]
\centering
\small
\tabcolsep=0.08cm
\begin{tabular}{p{0.15\textwidth} p{0.27\textwidth} p{0.27\textwidth} p{0.08\textwidth} p{0.08\textwidth} p{0.08\textwidth}}
\toprule[1.5pt]
\makecell{\textbf{Error}\\\textbf{Category}} & \makecell{\textbf{Examples}} & \makecell{\textbf{Possible}\\\textbf{Reasons}} & \makecell{\textbf{\# (\%)}\\\textbf{of All}\\\textbf{Err.}} & \makecell{\textbf{\# (\%)}\\\textbf{of Gem}\\\textbf{Err.}} & \makecell{\textbf{\# (\%)}\\\textbf{of Age}\\\textbf{Err.}} \\
\midrule[0.75pt]
Fail to render objects of the required number & Instruction: ``Create an image of an office desk that includes a stapler, a mouse, and a pen...'' Error: The generated image contains two pens, violating the ``exactly once'' constraint. & The model demonstrates limited numeracy and counting capabilities. This may be due to the attention mechanism, which often fails to disentangle identical object instances and leads to superfluous objects. & \errstat{12}{3.6\%} & \errstat{10}{4.2\%} & \errstat{2}{2.2\%} \\

Failed spatial generation: images are not in the position / angle asked & Instruction: ``Using the provided image as the reference angle, create four additional images...'' Error: The model generates two images at 30 degrees to the right and two at 30 degrees to the left. & Poor understanding of 3D position. Models are trained on image-caption pairs that stress semantic consistency and overlook other perception abilities. & \errstat{4}{1.2\%} & \errstat{2}{0.8\%} & \errstat{2}{2.2\%} \\

Failed temporal generation: images do not show the required adding / removing order & Instruction: ``Create three images that sequentially show the addition of a coffee mug, a notebook, and a pen...'' Error: The first image already shows the final result. & Agent models do not show strong planning capabilities on image generation tasks. ARMs also lack cross-image consistency mechanisms, so state is not maintained across steps. & \errstat{21}{6.4\%} & \errstat{20}{8.4\%} & \errstat{1}{1.1\%} \\

Failed logic generation: images do not show the required logical order & Instruction: ``Create three images, each featuring a single balloon of a different color... Arrange the images in alphabetical order...'' Error: The generated order is still red, yellow, and purple. & Agent models do not show strong planning capabilities on image generation tasks. ARMs also lack cross-image consistency mechanisms, so state is not maintained across steps. & \errstat{11}{3.3\%} & \errstat{7}{3.0\%} & \errstat{4}{4.3\%} \\

Wrong editing region & Instruction: ``Create an image after the utensils have been removed from the photo.'' Error: The model removes the food but leaves the utensils untouched. & Current training for image editing lacks verifiable and accurate reward, which makes the editing region inaccurate. & \errstat{5}{1.5\%} & \errstat{3}{1.3\%} & \errstat{2}{2.2\%} \\

Oversized / undersized editing region & Instruction: ``Create an image that displays the result after removing the man's wig...'' Error: The edit also removes part of the forehead and background. & Current training for image editing lacks verifiable and accurate reward, which makes the editing region inaccurate. & \errstat{2}{0.6\%} & \errstat{1}{0.4\%} & \errstat{1}{1.1\%} \\
\bottomrule[1.5pt]
\end{tabular}
\caption{Generation model failure analysis on interleaved image-text generation tasks (Part II). Analyzed models are \textsc{Gemini Image 2.0} and \textsc{Gemini 2.5 Pro} + \textsc{GPT Image} Agent.}
\label{tab:interleaved_failure_analysis_2}
\end{table*}
\begin{table*}[!ht]
\centering
\small
\tabcolsep=0.08cm
\begin{tabular}{p{0.15\textwidth} p{0.27\textwidth} p{0.27\textwidth} p{0.08\textwidth} p{0.08\textwidth} p{0.08\textwidth}}
\toprule[1.5pt]
\makecell{\textbf{Error}\\\textbf{Category}} & \makecell{\textbf{Examples}} & \makecell{\textbf{Possible}\\\textbf{Reasons}} & \makecell{\textbf{\# (\%)}\\\textbf{of All}\\\textbf{Err.}} & \makecell{\textbf{\# (\%)}\\\textbf{of Gem}\\\textbf{Err.}} & \makecell{\textbf{\# (\%)}\\\textbf{of Age}\\\textbf{Err.}} \\
\midrule[0.75pt]
Invalid edit: edit not applied to the required text / object & Instruction: ``Create an image that shows the result after replacing the cop with a smiling clown...'' Error: The output image is identical to the input image. & Under-training for image editing tasks. Models cannot understand the editing instruction well. & \errstat{20}{6.1\%} & \errstat{16}{6.8\%} & \errstat{4}{4.3\%} \\

Partial edit: edit only applies to a single object / text & Instruction: ``Generate an image showing the editing result after making all the sprinkles on the cupcakes blue...'' Error: Only the front cupcake is edited. & Under-training for image editing tasks. Models cannot understand the editing instruction well. & \errstat{4}{1.2\%} & \errstat{3}{1.3\%} & \errstat{1}{1.1\%} \\

Global changes outside the editing region & Instruction: ``Create an image displaying the result after replacing the highest kite with an eagle...'' Error: The eagle is added correctly, but the whole sky changes color. & Current training for image editing lacks verifiable and accurate reward, which makes the editing scope difficult to control. & \errstat{11}{3.3\%} & \errstat{4}{1.7\%} & \errstat{7}{7.6\%} \\

Generated image-text interleaved content not in the required modality order & Instruction: ``For each phase, start with an image that illustrates the phase, followed by a written explanation...'' Error: The model outputs all text first and all images at the end. & Lack of instruction-following training for interleaved image-text content. Models cannot maintain the required output structure and modality sequence. & \errstat{6}{1.8\%} & \errstat{5}{2.1\%} & \errstat{1}{1.1\%} \\

Parsing error: wrong function-calling format for image generation & Instruction: ``Develop a 3-step guide...'' Error: The model outputs raw tool code or no image at all. & Lack of training on agentic data. Models cannot reliably output the formatted text required by tool agents. & \errstat{0}{0.3\%} & \errstat{1}{0.4\%} & \errstat{0}{0.0\%} \\

Parsing error: failed to generate text in the required format & Instruction: ``...return only a JSON object that maps each item to its corresponding color...'' Error: The model returns a bulleted list instead of valid JSON. & Lack of training on agentic data. Models cannot reliably output the formatted text required by tool agents. & \errstat{19}{5.8\%} & \errstat{19}{8.0\%} & \errstat{0}{0.0\%} \\
\bottomrule[1.5pt]
\end{tabular}
\caption{Generation model failure analysis on interleaved image-text generation tasks (Part III). Analyzed models are \textsc{Gemini Image 2.0} and \textsc{Gemini 2.5 Pro} + \textsc{GPT Image} Agent.}
\label{tab:interleaved_failure_analysis_3}
\end{table*}
\begin{table*}[!ht]
\centering
\small
\tabcolsep=0.08cm
\begin{tabular}{p{0.15\textwidth} p{0.27\textwidth} p{0.27\textwidth} p{0.08\textwidth} p{0.08\textwidth} p{0.08\textwidth}}
\toprule[1.5pt]
\makecell{\textbf{Error}\\\textbf{Category}} & \makecell{\textbf{Examples}} & \makecell{\textbf{Possible}\\\textbf{Reasons}} & \makecell{\textbf{\# (\%)}\\\textbf{of All}\\\textbf{Err.}} & \makecell{\textbf{\# (\%)}\\\textbf{of Gem}\\\textbf{Err.}} & \makecell{\textbf{\# (\%)}\\\textbf{of Age}\\\textbf{Err.}} \\
\midrule[0.75pt]
Image inconsistent with the self-generated object description in text & Instruction: ``Create four images that sequentially show the result after removing the sunglasses, the camera, the map, and the passport...'' Error: The text says the passport is removed, but the final image still shows the passport. & Lack of training on interleaved image-text tasks. Models cannot produce self-consistent image-text pairs. For agent models, this is often because the image generation tool does not strictly follow the plan, and the planning model cannot correct it with feedback. & \errstat{23}{7.0\%} & \errstat{15}{6.3\%} & \errstat{8}{8.7\%} \\

Image inconsistent with the self-generated relation description in text & Instruction: ``...answer the following two questions...'' Error: The model answers ``Left,'' but the generated image places the object on the right. & Lack of training on interleaved image-text tasks. Models cannot produce self-consistent image-text pairs. For agent models, this is often because the image generation tool does not strictly follow the plan, and the planning model cannot correct it with feedback. & \errstat{45}{13.7\%} & \errstat{31}{13.1\%} & \errstat{14}{15.2\%} \\

Image inconsistent with the self-generated OCR recognition results in text & Instruction: ``...after generating the image, output only the announcement text in XML format...'' Error: The XML output says ``Science Fair,'' but the rendered image contains gibberish. & Lack of training on interleaved image-text tasks. Models cannot produce self-consistent image-text pairs. For agent models, this is often because the image generation tool does not strictly follow the plan, and the planning model cannot correct it with feedback. & \errstat{16}{4.9\%} & \errstat{14}{5.9\%} & \errstat{2}{2.2\%} \\

No step-by-step reasoning analysis & Instruction: ``Analyze the following SVG code step-by-step...'' Error: The model gives only a short summary instead of the required step-by-step reasoning. & Under-training for multimodal reasoning tasks involving image generation. RLVR may improve such multimodal reasoning capabilities if more verifiable data like \textsc{MMMG} is released. & \errstat{2}{0.6\%} & \errstat{2}{0.8\%} & \errstat{0}{0.0\%} \\

Wrong reasoning analysis & Instruction: ``What geometric shape does this SVG code describe?...'' Error: The model reasons incorrectly and concludes that the code describes a circle. & The model hallucinates the function of the code or cannot mentally simulate the SVG geometry. & \errstat{46}{14.0\%} & \errstat{22}{9.3\%} & \errstat{24}{26.1\%} \\

Failed generation & Instruction: ``...'' Error: The output is a completely black image or a corrupted file that cannot be opened. & Activation of safety filters or internal system generation failures. & \errstat{5}{1.5\%} & \errstat{4}{1.7\%} & \errstat{1}{1.1\%} \\
\bottomrule[1.5pt]
\end{tabular}
\caption{Generation model failure analysis on interleaved image-text generation tasks (Part IV). Analyzed models are \textsc{Gemini Image 2.0} and \textsc{Gemini 2.5 Pro} + \textsc{GPT Image} Agent.}
\label{tab:interleaved_failure_analysis_4}
\end{table*}

\end{document}